\documentclass[10pt,twocolumn,letterpaper]{article}

\usepackage{iccv}
\usepackage{times}
\usepackage{epsfig}
\usepackage{graphicx}
\usepackage{amsmath}
\usepackage{amssymb}
\usepackage{algorithm}
\usepackage{algorithmicx}
\usepackage{algpseudocode}
\usepackage{caption}
\usepackage{float} 
\usepackage{subcaption}
\usepackage{booktabs}
\usepackage{multirow}
\usepackage[accsupp]{axessibility}  

\usepackage[pagebackref=true,breaklinks=true,letterpaper=true,colorlinks,bookmarks=false]{hyperref}

\iccvfinalcopy 

\def\iccvPaperID{6592} 

\ificcvfinal\fi

\begin{document}

\title{GIFD: A Generative Gradient Inversion Method with Feature Domain Optimization}

\author{
Hao Fang$^{1,2}$
\quad
Bin Chen$^{ 1,3,4}$\thanks{Corresponding Author} 
\quad
Xuan Wang$^{1,3,4}$
\quad
Zhi Wang$^{2,3}$
\quad
Shu-Tao Xia$^{2}$\\
$^{1}$Harbin Institute of Technology, Shenzhen
\quad\\
$^{2}$Tsinghua Shenzhen International Graduate School, Tsinghua University\quad 
$^{3}$Peng Cheng Laboratory\\
$^{4}$Guangdong Provincial Key Laboratory of Novel Security Intelligence Technologies
\\
{\tt\small 190110304@stu.hit.edu.cn, chenbin2021@hit.edu.cn, wangxuan@cs.hit.edu.cn}
\\
{\tt\small \{wangzhi, xiast\}@sz.tsinghua.edu.cn}
}

\maketitle
\ificcvfinal\thispagestyle{empty}\fi

\def\thefootnote{*}
\def\thefootnote{\arabic{footnote}}
\maketitle
\ificcvfinal\thispagestyle{empty}\fi

\begin{abstract}

Federated Learning (FL) has recently emerged as a promising distributed machine learning framework to preserve clients' privacy, by allowing multiple clients to upload the gradients calculated from their local data to a central server. Recent studies find that the exchanged gradients also take the risk of privacy leakage, e.g., an attacker can invert the shared gradients and recover sensitive data against an FL system by leveraging pre-trained generative adversarial networks (GAN) as prior knowledge. However, performing gradient inversion attacks in the latent space of the GAN model limits their expression ability and generalizability. To tackle these challenges, we propose \textbf{G}radient \textbf{I}nversion over \textbf{F}eature \textbf{D}omains (GIFD), which disassembles the GAN model and searches the feature domains of the intermediate layers. Instead of optimizing only over the initial latent code, we progressively change the optimized layer, from the initial latent space to intermediate layers closer to the output images. In addition, we design a regularizer to avoid unreal image generation by adding a small ${l_1}$ ball constraint to the searching range.  We also extend GIFD to the out-of-distribution (OOD) setting, which weakens the assumption that the training sets of GANs and FL tasks obey the same data distribution. Extensive experiments demonstrate that our method can achieve pixel-level reconstruction and is superior to the existing methods. Notably, GIFD also shows great generalizability under different defense strategy settings and batch sizes.

\end{abstract}

\section{Introduction}
Federated learning \cite{mcmahan2017communication, zhang2021survey} is an increasingly popular distributed machine learning framework, which has been applied in many privacy-sensitive scenarios \cite{li2020review, yang2020federated}, such as financial services, medical analysis, and recommendation systems. 
\begin{figure}

    \begin{subfigure}{\linewidth}
        \begin{minipage}[t]{0.05\linewidth}
        \rotatebox{90}{{\textbf{~~ImageNet}}}   
        \end{minipage}%
        \begin{minipage}[t]{0.23\linewidth}
        \centering
        \includegraphics[width=1.7cm]{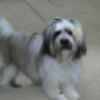}
        \centering

        \end{minipage}%
        \begin{minipage}[t]{0.23\linewidth}
        \centering
        \includegraphics[width=1.7cm]{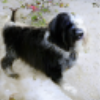}
        \centering

        \end{minipage}%
        \begin{minipage}[t]{0.23\linewidth}
        \centering
        \includegraphics[width=1.7cm]{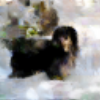}
        \centering

        \end{minipage}%
        \begin{minipage}[t]{0.23\linewidth}
        \centering
        \includegraphics[width=1.7cm]{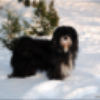}
        \centering

        \end{minipage}%
        \end{subfigure}
        
    \vskip 9pt
    
    \begin{subfigure}{\linewidth}
        \begin{minipage}[t]{0.05\linewidth}
        \rotatebox{90}{{\textbf{~~~~FFHQ}}}   
        \end{minipage}%
        \begin{minipage}[t]{0.23\linewidth}
        \centering
        \includegraphics[width=1.7cm]{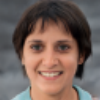}
        \centering
        \caption*{\textbf{\footnotesize{Dummy Input}}}
        \end{minipage}%
        \begin{minipage}[t]{0.23\linewidth}
        \centering
        \includegraphics[width=1.7cm]{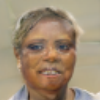}
        \centering
        \caption*{\textbf{\footnotesize{{Latent Space}}}}
        \end{minipage}%
        \begin{minipage}[t]{0.23\linewidth}
        \centering
        \includegraphics[width=1.7cm]{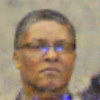}
        \centering
        \caption*{\textbf{\footnotesize{GIFD}}}
        \end{minipage}%
        \begin{minipage}[t]{0.23\linewidth}
        \centering
        \includegraphics[width=1.7cm]{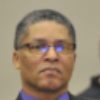}
        \centering
        \caption*{\textbf{\footnotesize{Ground Truth}}}
        \end{minipage}%
        \end{subfigure}

\caption{The reconstructed results of our proposed GIFD on ImageNet\cite{deng2009imagenet} and FFHQ\cite{karras2019style}. The first column contains the randomly initialized images generated by generators. The next two columns show the reconstruction samples of the latent space search and our proposed GIFD, } 
\label{contribution}
\end{figure}

It allows multiple clients to participate in collaborative learning under the coordination of the central server. The central server aggregates the uploaded gradients calculated from the local data by the end users, rather than the private data. This mechanism resolves the data silos problem and brings privacy benefits to distributed learning. However, a series of recent studies have shown that even the gradients uploaded in FL take the risk of privacy leakage. Zhu \etal \cite{zhu2019deep} first formulate it as an optimization problem and design an optimization-based algorithm that reconstructs private data by best matching the dummy gradients with the real gradients. Zhao \etal \cite{zhao2020idlg} further improve the attack with an extra label restoration step. Geiping \etal \cite{geiping2020inverting} first achieve ImageNet-level recovery through a well-designed loss function that adds a new regularization and uses a different distance metric. In order to improve the performance on larger batch sizes, Yin \etal \cite{yin2021see} propose a batch-level label extraction method and assume that certain side-information is available to regularize feature distributions through batch normalization (BN) prior.

It is widely investigated and acknowledged that a pre-trained GAN learned from a public dataset generally captures a wealth of prior knowledge. Recent studies \cite{yin2021see, jeon2021gradient, li2022auditing} propose to leverage the manifold of GAN as prior information, which provides a good approximation of the natural image space and enhances the attacks significantly. The aforementioned works achieve impressive results in their own scenarios, but most of them rely on strong assumptions, e.g., known labels, BN statistics, and private data distribution, which are actually impractical in the real FL scenario. Therefore, it is hard for most existing methods to recover high-quality private data in a more realistic setting.

In this paper, we advocate a simple and effective solution, Gradient Inversion over Feature Domain (GIFD), to address the challenges of expression ability and generalizability of pre-trained GANs. Recently, it has been shown that rich semantic information is encoded in the intermediate features and the latent space of GANs \cite{bau2019gan, tewari2020pie, shen2020interpreting,daras2021intermediate}. Among them, the GAN-based intermediate layer optimization in solving compressed sensing problems achieves great performance \cite{daras2021intermediate}. Inspired by these works, We reformulate the GAN inversion as a novel intermediate layer optimization problem by minimizing the gradient matching loss by searching the intermediate features of the generative model. Specifically, our first step is to optimize the latent space and then we optimize the intermediate layers of the generative model successively. During the feature domain optimization stage, we only use part of the generator and the solution space becomes larger, which can easily lead to unreal image generation. To solve this problem, we iteratively project the optimizing features to a small ${l_1}$ ball centered at the initial vector induced by the previous layer. Finally, we select output images from the layer with the corresponding least gradient matching loss as the final results. The visual comparison in Figure \ref{contribution}  clearly demonstrates the necessity of optimizing the intermediate feature domains.


Another issue unsolved in GAN-based gradient attacks is the flexibility of private data generation under more rigorous and realistic settings. To relax these assumptions, we first investigate an out-of-distribution (OOD) gradient attack scenario, where the private data distribution is significantly different from that of the GAN's training set. The significant result improvement demonstrates the proposed method has excellent generalizability and achieves great performance on OOD datasets. Furthermore, we discuss several common defense strategies in \textit{protection form gradient sharing}\cite{zhang2022survey}, including gradient sparsification \cite{Strom2015, aji-heafield-2017-sparse}, gradient clipping \cite{geyer2017differentially}, differential privacy  \cite{geyer2017differentially}, and Soteria (\ie, perturbing the data representations) \cite{sun2021soteria}. These frequently used privacy defense approaches have been confirmed to achieve high resilience against existing attacks by degrading the privacy information carried by the share gradients. Extensive experiments and ablation studies have demonstrated the effectiveness of the GIFD attack.



Our main contributions are summarized as follows:
\begin{itemize}
  \item We propose GIFD for exploiting pre-trained generative models as data prior to invert gradients by searching the latent space and the intermediate features of the generator successively with ${l_1}$ ball constraint.
  \item We show that this optimization method can be used to generate private OOD data with different styles, demonstrating the impressive generalization ability of the proposed GIFD under a more practical situation.
  \item We systematically evaluate our proposed method compared with the state-of-the-art baselines with the gradient transformation technique under four considered defense strategies.


\end{itemize}

\section{Related Work}

\subsection{Gradient-based Attack in FL}
In federated learning, the early studies investigate \emph{member inference} \cite{shokri2017membership,melis2019exploiting}, where a malicious attacker can determine whether a certain data sample has participated in model training. A similar attack, called \emph{property inference} \cite{ganju2018property}, can reveal the attributes of the samples in the training set. Another powerful attack is \emph{model inversion} \cite{hitaj2017deep}, which works by training a GAN from local images and the shared gradients to generate samples with the same distribution as the private data. Wang \etal \cite{wang2019beyond} then improve the model attack and reconstruct client-level data representatives.

\textbf{Gradient Inversion Attacks.} This is a more threatening type of 
attack where an adversary can fully reconstruct the client’s private data
samples. The existing attack methods can be characterized into 
two paradigms \cite{zhang2022survey}: \emph{recursion} and \emph{iteration}-based methods. 

Recursion-based attacks. Phong \etal \cite{8241854} first utilized gradients to successfully recover the input data from a shallow perceptron. Fan \etal \cite{fan2020rethinking} considered networks with convolution layers and solved the problem by converting the convolution layer into a full connection layer. Zhu \etal \cite{zhu2021rgap} combined forward and backward propagation to transform the problem into solving a system of linear equations. Chen \etal \cite{chen2021understanding} then combined optimization problems under different situations and proposed a systematic framework. The recursion-based methods have the following limitations: (1) low-resolution images only; (2) the global model in FL cannot contain pooling layers or shortcut connections; (3) these methods cannot handle mini-batch training; and (4) they heavily depend on gradients, \ie, if gradients are perturbed, most of these methods barely work.

Iteration-based attacks. Zhu \etal \cite{zhu2019deep} first formulated the attack as an iterative optimization problem. Attackers restore data samples by minimizing the distance between the shared gradients and the dummy gradients generated by a pair of dummy samples. Zhao \etal \cite{zhao2020idlg} proposed to extract the label of a single sample from the gradients and further improved the attack. Geiping \etal \cite{geiping2020inverting} reconstructed higher resolution images from ResNet \cite{he2016deep} by changing the distance metric and adding a regularization term. Yin \etal \cite{yin2021see} primarily focused on larger batch sizes recovery. With strong BN statistics and deep pre-trained ResNet-50 as the global model (larger model generates more gradient information), they successfully revealed some information from partial images at larger batch sizes. Jeon \etal \cite{jeon2021gradient} fine-tuned the GAN parameter to better utilize image prior and improved the quality of restored images. Hatamizadeh \etal \cite{hatamizadeh2022gradvit} extended attacks on Vision Transformers. Considering defense strategies in FL, Li \etal \cite{li2022auditing} proposed a new technique called gradient transformation to deal with the degraded gradients and still revealed private information.

Currently, several strong assumptions are made to help better reconstruct, which are not identical to the realistic FL setting. By nullifying some of these assumptions \cite{huang2021evaluating}, the reconstruction performance drops significantly.

\subsection{GAN as prior knowledge} 
GAN \cite{goodfellow2020generative} is a deep generative model, which can learn the probability distribution of the images in the training set through adversarial training. A well-trained GAN can generate realistic and high-diversity images. Recent studies show that GAN can be leveraged to solve inverse problems \cite{xia2022gan}, \eg compressed sensing. Yin \etal \cite{yin2021see} introduced a method that utilizes a pre-trained generative model as an image prior. Jeon \etal \cite{jeon2021gradient} proposed to search the latent space and parameter space of the generative model in turn, which fully exploits GAN's generation ability to reconstruct images of outstanding quality. A weakness is that it requires a specific generator to be trained for each reconstructed image, which consumes large amounts of GPU memory and inference time. Li \etal \cite{li2022auditing} also adopted the generative model, but only optimized the latent code, which achieves semantic-level reconstruction. Among the GAN-based methods, only Jeon \etal \cite{jeon2021gradient} really considered the situation when the training data of the generative model and the global model obey different probability distributions.
\begin{figure*}
\centering
\includegraphics[width=\linewidth]{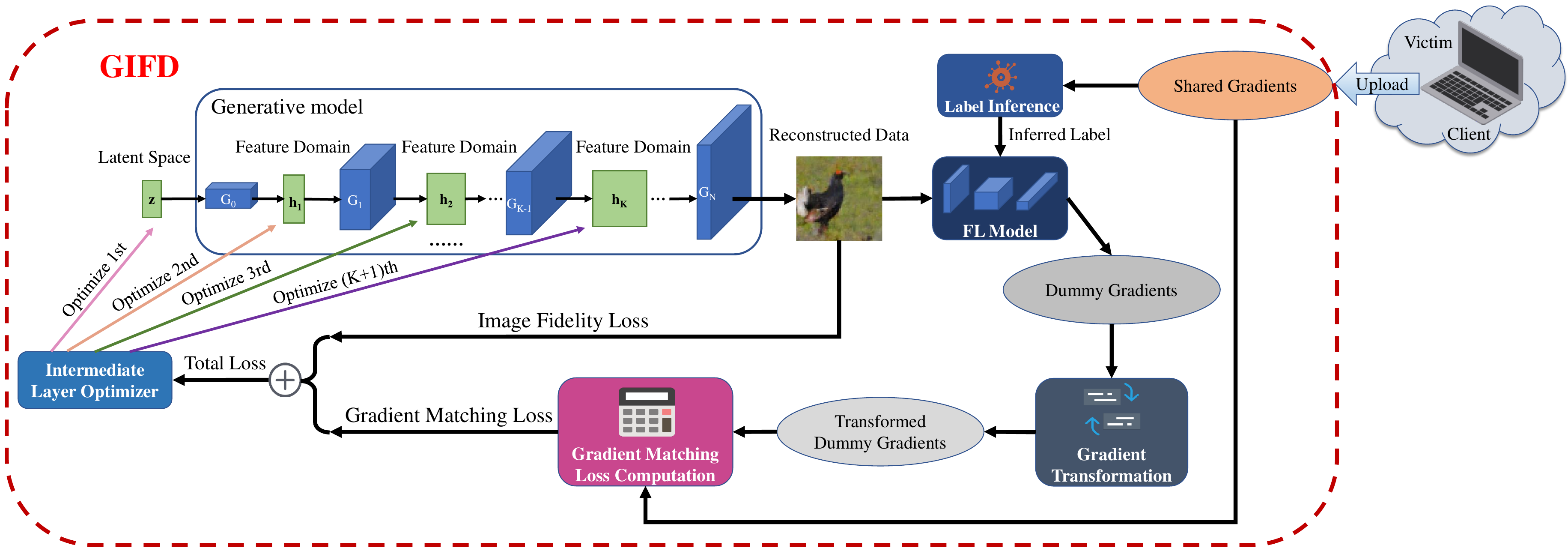}
\caption{Overview of our proposed GIFD attack. The intermediate layer optimizer minimizes the loss computed from the dummy gradients and the shared gradients from the victim under the image fidelity regularization, to update the latent vector and the intermediate features successively. Finally, the generative model outputs reconstruction data from the layer with the corresponding least gradient matching loss.} \label{pipeline_v5.pdf}
\end{figure*}

Inspired by the successful application of Intermediate Layer Optimization (ILO) \cite{daras2021intermediate} in compressed sensing, we decide to search the latent space and feature domains of the generative model to achieve pixel-level reconstruction. Meanwhile, we find that our method is superior to the previous methods for OOD data.

\section{Method}

In this section, we first introduce the basic paradigm of gradient inversion attacks. Then, we explain how former methods leverage GAN to achieve better results. Finally, we elaborate on our proposed GIFD, which successively searches the latent space and intermediate feature spaces of the generative model.

\subsection{Problem Formulation}
Given a neural network $f_\theta$ with weights $\theta$ for image classification tasks, and batch-averaged gradients $g$ calculated from a private batch with images $\mathbf{x^*}$ and labels $\mathbf{y^*}$, the attacker attempts to invert the gradients to private data with randomly initialized input tensor $\mathbf{\hat{x}} \in \mathbb{R}^{B\times H\times W \times C}$ and labels $\mathbf{\hat{y}} \in \{0,1\}^{B\times L}$ ($B,H,W,C,L$ being batch size, height, width, number of channels and class number):
\begin{equation}
\begin{aligned}
    \mathbf{\hat{x}^*}, \mathbf{\hat{y}^*} = \mathop{\arg\min}_{\mathbf{\mathbf{\hat{x}}}, \mathbf{\hat{y}}} \mathcal{D}\left(\frac{1}{B}\sum_{i=1}^{B}\nabla\ell(f_\theta(x_i),y_i), g\right),
\end{aligned}
\end{equation}
where $\mathbf{\hat{x}}=(x_1,\dots, x_B)$, $\mathbf{\hat{y}}=(y_1,\dots, y_B)$. $\mathcal{D}(\cdot,\cdot)$ is the measurement of distance, \eg, $l_2$-distance \cite{yin2021see,li2022auditing}, negative cosine similarity \cite{geiping2020inverting, jeon2021gradient}, and $\ell(\cdot,\cdot)$ is the loss function for classification. In the workflow of the algorithm, the attacker generates a pair of random noise $\mathbf{\hat{x}}$ and labels $\mathbf{\hat{y}}$ as parameters, optimized towards the ground truth $\mathbf{x^*}$ and $\mathbf{y^*}$ through minimizing the matching loss between dummy gradients and transmitted gradients. 


Since private labels can be inferred directly from the gradients \cite{zhao2020idlg, yin2021see}, the objective function with regularization term can be simplified to the following form:

\begin{equation}
\begin{aligned}
\mathbf{\hat{x}^*}= \mathop{\arg\min}_{\mathbf{\mathbf{\hat{x}}}}\mathcal{D}\left(F(\mathbf{\hat{x}}), g\right) + R_{prior}(\mathbf{\hat{x}}),
\end{aligned}
\end{equation}
where $F(\mathbf{\hat{x}})=\frac{1}{B}\sum_{i=1}^{B}\nabla\ell(f_\theta(x_i),y_i)$, $R_{prior}(\mathbf{\hat{x}})$ is prior knowledge regularization (\eg, BN statistics \cite{yin2021see}).

Given a pre-trained generative model $G_w(\cdot)$ learning from the public dataset, an intuitive method is to transform the problem into the following form:

\begin{equation}
\begin{aligned}
\mathbf{z^*}= \mathop{\arg\min}_{\mathbf{\mathbf{z}}}\mathcal{D}\left(F(G_w(\mathbf{z})), g\right) + R_{prior}(\mathbf{z};G_w),
\end{aligned}
\end{equation}
where $\mathbf{z} \in \mathbb{R}^{B\times k}$ is the latent code of the generative model. By narrowing the search range from $\mathbb{R}^{B\times m}$ ($m=H\times W \times C$) to $\mathbb{R}^{B\times k}$ ($k \textless\textless m$), one can reduce the uncertainty in the optimizing process. Based on this, various GAN-based gradient inversion methods \cite{li2022auditing,jeon2021gradient} are proposed to ensure the quality and fidelity of the generated images.


\subsection{Gradient Inversion over Feature Domains}
First, we formally formulate our optimization objective:
\begin{equation} \label{(4)}
\begin{aligned}
\mathbf{\hat{x}^*}= \mathop{\arg\min}_{\mathbf{\hat{x}}}\mathcal{D}\left(\mathcal{T}(F(\mathbf{\hat{x}})), g\right) + \mathcal{R}_{fidty}(\mathbf{\hat{x}}),
\end{aligned}
\end{equation}
where $\mathbf{\hat{x}}$ is generated by $G_w$ or part of $G_w$, $F(\cdot)$ is the batch-averaged gradient operator, $\mathcal{T}(\cdot)$ is the gradient transformation technique we will discuss later. The first term $\mathcal{D}\left(\mathcal{T}(F(\mathbf{\hat{x}})), g\right)$ denotes the gradient matching loss, and the second term $\mathcal{R}_{fidty}(\mathbf{\hat{x}})$ is the image fidelity regularization. To simplify the expression, we solve for the objective function in the following form:

\begin{equation}
\begin{aligned}
\mathbf{\hat{x}^*} = \mathop{\arg\min}_{\mathbf{\hat{x}}} \mathcal{L}_{grad}(\mathbf{\hat{x}}),
\end{aligned}
\end{equation}
where we denote the loss function in (\ref{(4)}) by $ \mathcal{L}_{grad}(\mathbf{\hat{x}})$. An overview of our method is shown in Figure \ref{pipeline_v5.pdf}, we next introduce each component in detail.

\noindent\textbf{Intermediate Layer Optimizer.} This is the core of our algorithm. As the pseudocode described in Algorithm \ref{GIFD}, instead of directly optimizing over $\mathbf{\hat{x}}$, we focus on searching the latent space and the intermediate space of the generator in turn, to make the most of the GAN prior. 

The first step is to optimize over the randomly initialed latent vector $\mathbf{z}$ using gradient descent with an effective Spherical Optimizer \cite{menon2020pulse}. Once we obtain the optimal $\mathbf{z^*}$, we dissemble the generator $G_w$ into $G_0\circ G_1\circ \dots \circ G_{N-1} \circ G_N$ for intermediate feature optimization. Then, we map optimal latent vector $\mathbf{z^*}$ into intermediate latent representations $\mathbf{h_1^0}$ using $G_0$, \ie, $\mathbf{h_1^0}:=G_0(\mathbf{z^*})$. Next, our algorithm enters the for loop in line \ref{line7} of Algorithm \ref{GIFD} and starts to search the intermediate features.

At the pass of loop $i$, we perform the following operations. First, we generate images from intermediate feature $\mathbf{h_i}$ only with the rest part of $G_w$ (\ie, $G_i\circ \dots \circ G_N$). Then, we use the generated images to compute dummy gradients and optimize over $\mathbf{h_i}$ via minimizing cost function in (\ref{(4)}). Considering the intermediate feature searching might lead to unreal images generation, we constrain the searching range to lie within an $l_1$ ball of radius $r[i]$ centered at $\mathbf{{h_i^0}}$, \ie the term $ball_{\mathbf{h_i^0}}^{r[i]}$ in the line \ref{line9} of Algorithm \ref{GIFD}. After obtaining the optimal results $\mathbf{h_i^*}$ of the present layer, we generate the initial intermediate representations for the next layer with $G_i$, \ie $\mathbf{h_{i+1}^0}:=G_i(\mathbf{h_i^*})$.

As shown in line \ref{line4}, \ref{line11}, \ref{line12}, \ref{line13}, \ref{line18} of Algorithm \ref{GIFD}, we hope to utilize the gradient matching loss as valid information to guide us to select the output images. More specifically, we choose the output images from the layer with the corresponding least gradient matching loss among all the searched intermediate layers as the final output. Although less loss doesn't always mean better image quality, our strategy still outperforms specifying a fixed layer's output. 

\renewcommand{\algorithmicrequire}{\textbf{Input:}}  
\renewcommand{\algorithmicensure}{\textbf{Output:}} 

\begin{algorithm}[h]
  \caption{Pseudocode of our proposed GIFD} \label{GIFD}
  \begin{algorithmic}[1]
    \Require
      $G_w$: a pre-trained generative model;
      $f_\theta$: the global model in FL;
      $g$: shared gradients;
      $K$: the index of the last intermediate layer to optimize;
      $r[1\dots K]$: radius of ${l_1}$ ball in each intermediate layer;
      $B$: batch size;

    \Ensure
       Reconstructed images via GIFD attack;
       \State Initial latent code $\mathbf{z}:=(z_1, \dots , z_B)$ with random noise
            \State \parbox[t]{\dimexpr\linewidth-\algorithmicindent}{\texttt{// Latent space search\strut}}
       \State $\mathbf{z^*} \leftarrow \mathop{\arg\min}_{\mathbf{z}} \mathcal{L}_{grad}(G_w(\mathbf{z}))$ 
       \State Set $\mathbf{\hat{x}^*}:=G_w(\mathbf{z^*})$, $loss_{min} = \mathcal{D}\left(\mathcal{T}(F(G_w(\mathbf{z^*}))), g\right)$   \label{line4}
       \State Dissemble $G_w$ into $G_0\circ G_1\circ \dots \circ G_{N-1} \circ G_N$
       \State Set $\mathbf{h_1^0} := G_0(\mathbf{z^*})$ 
       \For{$i \leftarrow 1$ to $K$} \label{line7}
            
            \State \parbox[t]{\dimexpr\linewidth-\algorithmicindent}{\texttt{//Intermediate layers search with $l_1$-ball constraint\strut}}

        \State $\mathbf{h_i^*} \leftarrow argmin_{\mathbf{h_i}\in  ball_{\mathbf{h_i^0}}^{r[i]}}\mathcal{L}_{grad}(G_i\circ \dots \circ G_N(\mathbf{h_i})) $  \label{line9}
        \State $loss_i = \mathcal{D}\left(\mathcal{T}(F(G_i\circ \dots \circ G_N(\mathbf{h_i^*}))), g\right)$
      \If {$loss_i<loss_{min}$}  \label{line11} \State $\mathbf{\hat{x}^*} := G_i\circ \dots \circ G_N(\mathbf{h_i^*})$  \label{line12}
      \State $loss_{min}=loss_i$  \label{line13}
     \EndIf
     \State \parbox[t]{\dimexpr\linewidth-\algorithmicindent}{\texttt{// Generate features of the next intermediate layer as the initial vector to optimize\strut}}

       \State $\mathbf{h_{i+1}^0} := G_i(\mathbf{h_i^*}) $
     \EndFor
  \State Return results: $\mathbf{\hat{x}^*}$ \label{line18}
  \end{algorithmic}
\end{algorithm}

With all the efforts above, we encourage the optimizer to explore the intermediate space with rich information, to generate more diverse and high-fidelity images, while limiting the solution space within a $l_1$ ball around the manifold induced by the previous layer in order to avoid overfitting and guarantee the realism of the generated images. Furthermore, our approach is easy to implement as it is not tied to any specific GAN architecture and only requires a pre-trained generative model.


\noindent\textbf{Labels Extraction.} Specifically, consider a network parameterized by $\rm W$ for classification task over $n$-classes using cross-entropy loss function, when the training data is a single image, the ground truth label $c$ can be accurately inferred \cite{zhao2020idlg} through:
\begin{equation}
\begin{aligned}
 c = i, \ \ \ {\rm s.t.} \ \nabla \mathbf{{\rm W}_{FC}^i{^{\top}}} \cdot \nabla \mathbf{{\rm W}_{FC}^j} \leq 0, \ \forall\ j \neq i,
\end{aligned}
\end{equation}
where we denote the gradient vector \wrt the weights (denoted as $\mathbf{{\rm W}_{FC}^i}$) connected to the $i_{th}$ logit in the classification layer (\ie, the output layer) by $\mathbf{\nabla {\rm W}_{FC}^i}$. Hence, we can identify the ground-truth label via the index of the negative gradients. \cite{yin2021see} further extends to support batch-level label extraction with high accuracy, while assuming non-repeating labels in the batch. The inferred labels are used to compute dummy gradients and as the class conditions for conditional GANs, which greatly enhances our attack.

\noindent\textbf{Image Fidelity Regularization.} Intuitively, it is challenging to restore data only from the shared gradients, as gradients are only a non-linear mapping form of the original data. It is therefore worth using some strong priors as an approximation of natural images:
\begin{equation}
\begin{aligned}
\mathcal{R}_{fidty}(\mathbf{\hat{x}}) = \alpha_{\ell_2}\mathcal{R}_{\ell_2}(\mathbf{\hat{x}})+\alpha_{TV}\mathcal{R}_{TV}(\mathbf{\hat{x}}),
\end{aligned}
\end{equation}
where the first term is the $l_2$ norm of the images \cite{yin2021see} with scaling factor $\alpha_{\ell_2}$, which encourages the algorithm to solve for a solution that is preferably sparse. Since neighboring pixels of natural images are likely to have close values, we add the second term \cite{geiping2020inverting} $\mathcal{R}_{TV}(\mathbf{\hat{x}})$ to penalize total variation of $\mathbf{\hat{x}}$ with scaling factor $\alpha_{TV}$.

\noindent\textbf{Gradient Transformation.} In order to mitigate the effects of defense strategies,  we adopt the adaptive attack \cite{li2022auditing} by estimating transformation from received gradients and incorporating it into the optimization process, \ie, $\mathcal{T}(\cdot)$ in (\ref{(4)}). Specifically, we can infer three defense strategies: (1) \textit{Gradient clipping}; (2) \textit{Gradient sparsification}; and (3) \textit{Soteria}.

\begin{figure}
	\centering
	\begin{subfigure}{0.495\linewidth}
		\centering
		\includegraphics[width=0.9\linewidth]{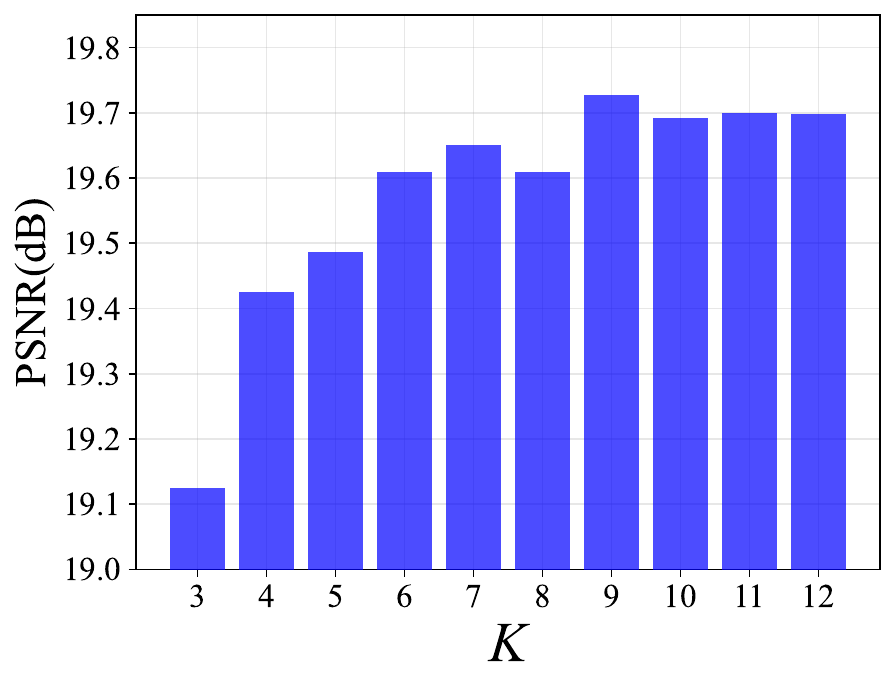}
		\caption{BigGAN}
		\label{select layer}
	\end{subfigure}
	\centering
	\begin{subfigure}{0.495\linewidth}
		\centering
		\includegraphics[width=0.9\linewidth]{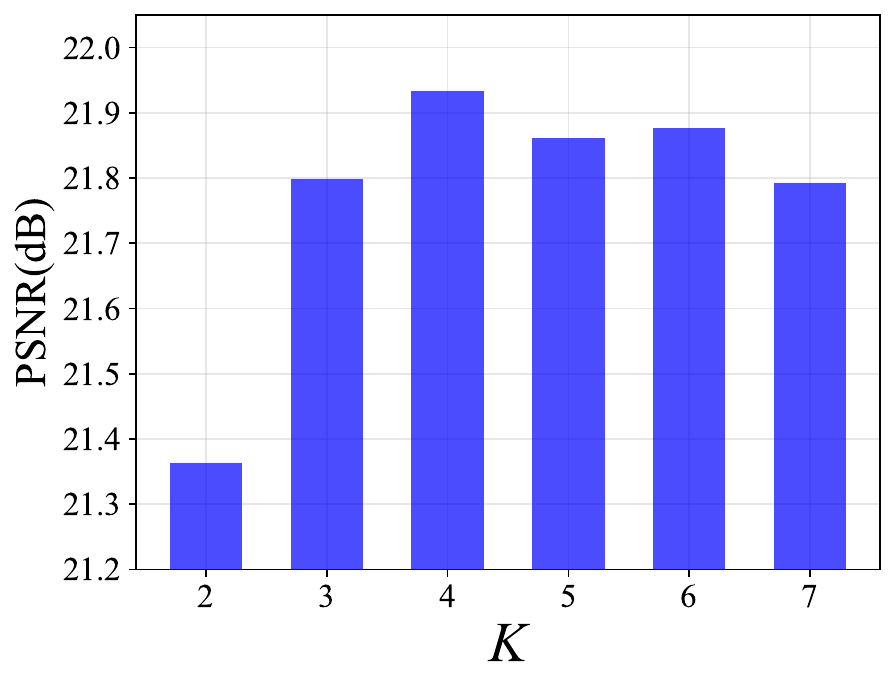}
		\caption{StyleGAN2}
		\label{select layer ffhq}
	\end{subfigure}
	\centering
	\caption{Comparison of PSNR mean on BigGAN and StyleGAN2 under different values of hyper-parameter $K$ (\ie, the last intermediate layer to optimize). Notably, the figures exclude the results where the corresponding values are below the starting point of the y-axis.}
    \vspace{-0.4cm}
	\label{select_layer_psnr}
\end{figure}

\section{Experiments}
\label{noise}
To validate the effectiveness of GIFD in improving attack performance, we conduct experiments on two widely used GANs in a range of scenarios. We evaluate our method for the classification task on the validation set of ImageNet ILSVRC 2012 dataset\cite{deng2009imagenet}) and 10-class (using age as label) FFHQ \cite{karras2019style} at $64\times64$ pixels. For the generative model, we use a pre-trained BigGAN \cite{brock2018large} for ImageNet and a pre-trained StyleGAN2 \cite{karras2019style} for FFHQ. We use a randomly initialized ResNet-18 as the FL model, and choose negative cosine similarity as distance metric $\mathcal{D}(\cdot)$. We use the default $B=1$ at one local step. Then we conduct experiments with larger $B$ and compare the performance of different methods. Our code is available at \textcolor{magenta}{\url{https://github.com/ffhibnese/GIFD}}.

\noindent\textbf{Implementation details}. According to its specific structure, we split BigGAN into $G_0$ to $G_{12}$ with 12 intermediate feature domains, and StyleGAN2 into $G_0$ to $G_7$ with 7 intermediate feature domains. We ensure that the intermediate features lie in the $l_1$ ball through Project Gradient Descent (PGD) \cite{nesterov2003introductory}. Motivated by the fact that a stepwise optimization over the noises in StyleGAN2 yields better reconstructions \cite{daras2021intermediate} for compressed sensing, we gradually allow to optimize more noises as we move to deeper intermediate layers and make them lie inside the $l_1$ ball as well. For more details about experiments, please refer to the Appendix.

\begin{table*}[htbp]
  \centering
  \caption{Comparison of GIFD with state-of-the-art methods on every 1000th image of the ImageNet and FFHQ validation set. We calculate the average value of metrics on reconstructed images.}
    \resizebox{15cm}{!}{\begin{tabular}{ccccccccccc} \toprule
    \multicolumn{1}{c}{\multirow{2}[0]{*}{Metric}} & \multicolumn{5}{c}{ImageNet}          & \multicolumn{5}{c}{FFHQ} \\ \cmidrule(lr){2-6} \cmidrule(lr){7-11} 
    \multicolumn{1}{c}{} & IG \cite{geiping2020inverting}    & GI \cite{yin2021see}    & GGL \cite{li2022auditing}   & GIAS \cite{jeon2021gradient}  & \textbf{GIFD}  & \multicolumn{1}{l}{IG \cite{geiping2020inverting}} & \multicolumn{1}{l}{GI \cite{yin2021see}} & \multicolumn{1}{l}{GGL \cite{li2022auditing}} & \multicolumn{1}{l}{GIAS \cite{jeon2021gradient}} & \multicolumn{1}{l}{\textbf{GIFD}} \\ \midrule
    PSNR$\uparrow$  & 17.0756  & 16.5109  & 13.3885  & 17.4923  & \textbf{20.0534 } & 15.3523  & 14.9485  & 15.1335  & 20.1799  & \textbf{21.3368} \\
    LPIPS$\downarrow$ & 0.3078  & 0.3297  & 0.3678  & 0.2536  & \textbf{0.1559 } & 0.4172  & 0.4503  & 0.2009  & 0.1266  & \textbf{0.1023 } \\
    SSIM$\uparrow$  & 0.2908  & 0.2673  & 0.1251  & 0.3381  & \textbf{0.4713 } & 0.2272  & 0.2044  & 0.2453  & 0.5379  & \textbf{0.5768} \\
    MSE$\downarrow$  & 0.0223  & 0.0258  & 0.0553  & 0.0236  & \textbf{0.0141 } & 0.0311  & 0.0343  & 0.0339  & 0.0121  & \textbf{0.0098} \\  \bottomrule
    \end{tabular}}
  \label{main_tab}%
\end{table*}

\begin{figure*}[htbp]
	\centering
	\begin{subfigure}{0.495\linewidth}
		\begin{minipage}[t]{0.165\linewidth}
        \centering
        \includegraphics[width=1.4cm]{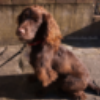}
        \end{minipage}%
        \begin{minipage}[t]{0.165\linewidth}
        \centering
        \includegraphics[width=1.4cm]{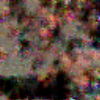}
        \end{minipage}%
        \begin{minipage}[t]{0.165\linewidth}
        \centering
        \includegraphics[width=1.4cm]{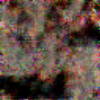}
        \end{minipage}%
        \begin{minipage}[t]{0.165\linewidth}
        \centering
        \includegraphics[width=1.4cm]{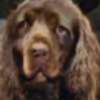}
        \end{minipage}%
        \begin{minipage}[t]{0.165\linewidth}
        \centering
        \includegraphics[width=1.4cm]{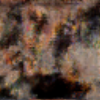}
        \end{minipage}%
        \begin{minipage}[t]{0.165\linewidth}
        \centering
        \includegraphics[width=1.4cm]{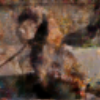}
        \end{minipage}%
        \qquad 
        \begin{minipage}[t]{0.165\linewidth}
        \centering
        \includegraphics[width=1.4cm]{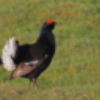}
        \end{minipage}%
        \begin{minipage}[t]{0.165\linewidth}
        \centering
        \includegraphics[width=1.4cm]{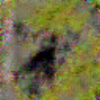}
        \end{minipage}%
        \begin{minipage}[t]{0.165\linewidth}
        \centering
        \includegraphics[width=1.4cm]{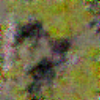}
        \end{minipage}%
        \begin{minipage}[t]{0.165\linewidth}
        \centering
        \includegraphics[width=1.4cm]{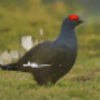}
        \end{minipage}%
        \begin{minipage}[t]{0.165\linewidth}
        \centering
        \includegraphics[width=1.4cm]{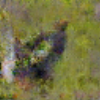}
        \end{minipage}%
        \begin{minipage}[t]{0.165\linewidth}
        \centering
        \includegraphics[width=1.4cm]{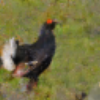}
        \end{minipage}
        \qquad 
        \begin{minipage}[t]{0.165\linewidth}
        \centering
        \includegraphics[width=1.4cm]{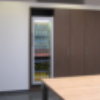}
        \end{minipage}%
        \begin{minipage}[t]{0.165\linewidth}
        \centering
        \includegraphics[width=1.4cm]{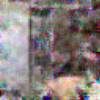}
        \end{minipage}%
        \begin{minipage}[t]{0.165\linewidth}
        \centering
        \includegraphics[width=1.4cm]{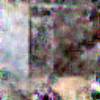}
        \end{minipage}%
        \begin{minipage}[t]{0.165\linewidth}
        \centering
        \includegraphics[width=1.4cm]{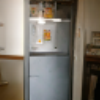}
        \end{minipage}%
        \begin{minipage}[t]{0.165\linewidth}
        \centering
        \includegraphics[width=1.4cm]{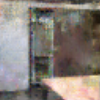}
        \end{minipage}%
        \begin{minipage}[t]{0.165\linewidth}
        \centering
        \includegraphics[width=1.4cm]{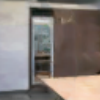}
        \end{minipage}
                \qquad 
        \begin{minipage}[t]{0.165\linewidth}
        \centering
        \includegraphics[width=1.4cm]{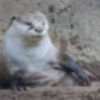}
        \centering
        \caption*{\textbf{\footnotesize{Original}}}
        \end{minipage}%
        \begin{minipage}[t]{0.165\linewidth}
        \centering
        \includegraphics[width=1.4cm]{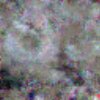}
        \centering
        \caption*{\textbf{\footnotesize{IG \cite{geiping2020inverting}}}}
        \end{minipage}%
        \begin{minipage}[t]{0.165\linewidth}
        \centering
        \includegraphics[width=1.4cm]{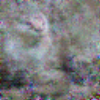}
        \centering
        \caption*{\textbf{\footnotesize{GI \cite{yin2021see}}}}
        \end{minipage}%
        \begin{minipage}[t]{0.165\linewidth}
        \centering
        \includegraphics[width=1.4cm]{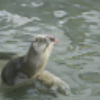}
        \centering
        \caption*{\textbf{\footnotesize{GGL \cite{li2022auditing}}}}
        \end{minipage}%
        \begin{minipage}[t]{0.165\linewidth}
        \centering
        \includegraphics[width=1.4cm]{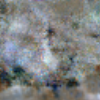}
        \centering
        \caption*{\textbf{\footnotesize{GIAS \cite{jeon2021gradient}}}}
        \end{minipage}%
        \begin{minipage}[t]{0.165\linewidth}
        \centering
        \includegraphics[width=1.4cm]{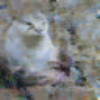}
        \centering
        \caption*{\textbf{\footnotesize{GIFD}}}
        \end{minipage}
	
	\caption{ImageNet (BigGAN)}
        \end{subfigure}
	\centering
	\begin{subfigure}{0.495\linewidth}
				\begin{minipage}[t]{0.165\linewidth}
        \centering
        \includegraphics[width=1.4cm]{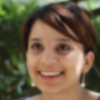}
        \end{minipage}%
        \begin{minipage}[t]{0.165\linewidth}
        \centering
        \includegraphics[width=1.4cm]{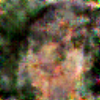}
        \end{minipage}%
        \begin{minipage}[t]{0.165\linewidth}
        \centering
        \includegraphics[width=1.4cm]{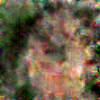}
        \end{minipage}%
        \begin{minipage}[t]{0.165\linewidth}
        \centering
        \includegraphics[width=1.4cm]{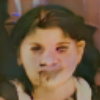}
        \end{minipage}%
        \begin{minipage}[t]{0.165\linewidth}
        \centering
        \includegraphics[width=1.4cm]{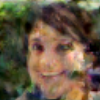}
        \end{minipage}%
        \begin{minipage}[t]{0.165\linewidth}
        \centering
        \includegraphics[width=1.4cm]{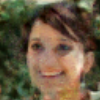}
        \end{minipage}%
        \qquad 
        \begin{minipage}[t]{0.165\linewidth}
        \centering
        \includegraphics[width=1.4cm]{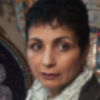}
        \end{minipage}%
        \begin{minipage}[t]{0.165\linewidth}
        \centering
        \includegraphics[width=1.4cm]{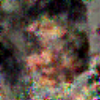}
        \end{minipage}%
        \begin{minipage}[t]{0.165\linewidth}
        \centering
        \includegraphics[width=1.4cm]{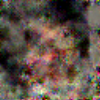}
        \end{minipage}%
        \begin{minipage}[t]{0.165\linewidth}
        \centering
        \includegraphics[width=1.4cm]{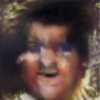}
        \end{minipage}%
        \begin{minipage}[t]{0.165\linewidth}
        \centering
        \includegraphics[width=1.4cm]{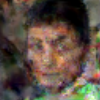}
        \end{minipage}%
        \begin{minipage}[t]{0.165\linewidth}
        \centering
        \includegraphics[width=1.4cm]{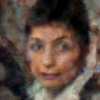}
        \end{minipage}
        \qquad 
        \begin{minipage}[t]{0.165\linewidth}
        \centering
        \includegraphics[width=1.4cm]{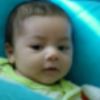}
        \end{minipage}%
        \begin{minipage}[t]{0.165\linewidth}
        \centering
        \includegraphics[width=1.4cm]{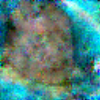}
        \end{minipage}%
        \begin{minipage}[t]{0.165\linewidth}
        \centering
        \includegraphics[width=1.4cm]{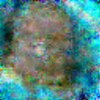}
        \end{minipage}%
        \begin{minipage}[t]{0.165\linewidth}
        \centering
        \includegraphics[width=1.4cm]{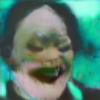}
        \end{minipage}%
        \begin{minipage}[t]{0.165\linewidth}
        \centering
        \includegraphics[width=1.4cm]{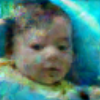}
        \end{minipage}%
        \begin{minipage}[t]{0.165\linewidth}
        \centering
        \includegraphics[width=1.4cm]{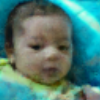}
        \end{minipage}
                \qquad 
        \begin{minipage}[t]{0.165\linewidth}
        \centering
        \includegraphics[width=1.4cm]{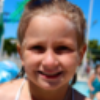}
        \centering
        \caption*{\textbf{\footnotesize{Original}}}
        \end{minipage}%
        \begin{minipage}[t]{0.165\linewidth}
        \centering
        \includegraphics[width=1.4cm]{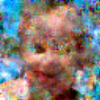}
        \centering
        \caption*{\textbf{\footnotesize{IG \cite{geiping2020inverting}}}}
        \end{minipage}%
        \begin{minipage}[t]{0.165\linewidth}
        \centering
        \includegraphics[width=1.4cm]{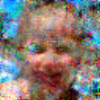}
        \centering
        \caption*{\textbf{\footnotesize{GI \cite{yin2021see}}}}
        \end{minipage}%
        \begin{minipage}[t]{0.165\linewidth}
        \centering
        \includegraphics[width=1.4cm]{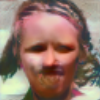}
        \centering
        \caption*{\textbf{\footnotesize{GGL \cite{li2022auditing}}}}
        \end{minipage}%
        \begin{minipage}[t]{0.165\linewidth}
        \centering
        \includegraphics[width=1.4cm]{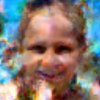}
                \centering
        \caption*{\textbf{\footnotesize{GIAS \cite{jeon2021gradient}}}}
        \end{minipage}%
        \begin{minipage}[t]{0.165\linewidth}

        \includegraphics[width=1.4cm]{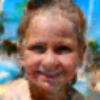}
        \centering
        \caption*{\textbf{\footnotesize{GIFD}}}
        \end{minipage}
	
            \caption{FFHQ (StyleGAN2)}
        \end{subfigure}
	\centering

	\caption{Qualitative results of different methods on ImageNet and FFHQ. }
	\label{main_figure}
\end{figure*}

\noindent\textbf{Evaluaion Metrics}. We compute the following quantitative metrics to measure the discrepancy between reconstructed images and ground truth: (1) PSNR (Peak Signal-to-Noise Ratio), (2) LPIPS \cite{zhang2018unreasonable} (Learned Perceptual Image Patch Similarity), (3) SSIM (Similarity Structural Index Measure), and (4) MSE (Mean Square Error) between reconstruction and private images.

\subsection{Decide Which Layer to End}
In order to further improve the quality of output images, we need to carefully handle the parameter $K$ in Algorithm \ref{GIFD}. Actually, we find that there is a trade-off between under-fitting and over-fitting about the choice of $K$. When $K$ is small, we only search the first few intermediate features of the generative model and do not fully utilize the rich information encoded in the intermediate space. As a result, the quality of the generated images does not meet our expectations. On the contrary, when $K$ is large, we excessively search the deeper layers and generate images that have less cost, but a larger discrepancy with the original images. Therefore, we randomly select images (disjoint from our main experimental data) from the validation set of ImageNet and FFHQ to study the impact of $K$ and try to select the best final layer. As shown in Figure \ref{select_layer_psnr}, when $K=9$ and $K=4$ are used for BigGAN and StyleGAN2 respectively, we obtain results with the largest PSNR. Hence, we use this configuration for conducting all the experiments.


\subsection{Comparison with the State-of-the-art Attacks}

Next, we compare our proposed GIFD with existing methods and provide qualitative and quantitative results. We consider the following four state-of-the-art baselines: (1) \textit{Inverting Gradients (IG)} by Geiping \etal  \cite{geiping2020inverting}; (2) \textit{GradInversion (GI)} by Yin \etal  \cite{yin2021see}; (3) \textit{Gradient Inversion in Alternative Spaces (GIAS)} by Jeon \etal  \cite{jeon2021gradient}; and (4) \textit{Generative Gradient Leakage (GGL)} by Li \etal  \cite{li2022auditing}.

In real application scenarios, a vast majority of FL systems do not transmit the BN statistics computed from private data \cite{huang2021evaluating}. Based on this fact, all the experiments do not use the strong BN prior proposed by  \cite{yin2021see}. Since the randomly initialized values of vectors will greatly affect the reconstruction results, we conduct 4 trials for every attack and select the result with the least gradient matching loss. The ablation study is conducted in the Appendix.

\begin{table*}[htbp]
  \centering
  \caption{Comparision of GIFD with state-of-the-art baselines on OOD data of different styles.}
    \resizebox{\linewidth}{!}{\begin{tabular}{ccllllllllllll}\toprule
    \multirow{2}[0]{*}{Datset} & \multicolumn{1}{c}{\multirow{2}[0]{*}{Method}} & \multicolumn{4}{c}{Art Painting} & \multicolumn{4}{c}{Photo}     & \multicolumn{4}{c}{Cartoon} \\ \cmidrule(lr){3-6} \cmidrule(lr){7-10} \cmidrule(lr){11-14}
          &       & \multicolumn{1}{c}{PSNR$\uparrow$} & \multicolumn{1}{c}{LPIPS$\downarrow$} & \multicolumn{1}{c}{SSIM$\uparrow$} & \multicolumn{1}{c}{MSE$\downarrow$} & \multicolumn{1}{c}{PSNR$\uparrow$} & \multicolumn{1}{c}{LPIPS$\downarrow$} & \multicolumn{1}{c}{SSIM$\uparrow$} & \multicolumn{1}{c}{MSE$\downarrow$} & \multicolumn{1}{c}{PSNR$\uparrow$} & \multicolumn{1}{c}{LPIPS$\downarrow$} & \multicolumn{1}{c}{SSIM$\uparrow$} & \multicolumn{1}{c}{MSE$\downarrow$} \\ \midrule
    \multirow{5}[0]{*}{ImageNet*} & IG \cite{geiping2020inverting}    & 18.3476  & 0.2286  & 0.3870  & 0.0172  & 15.6647  & 0.3575  & 0.2409  & 0.0325  & 15.8766  & 0.3183  & 0.3970  & 0.0288  \\
          & GI \cite{yin2021see}    & 17.4681  & 0.2625  & 0.3445  & 0.0203  & 15.2700  & 0.3888  & 0.2201  & 0.0346  & 15.3905  & 0.3112  & 0.3926  & 0.0327  \\
          & GGL \cite{li2022auditing}   & 12.8011  & 0.3639  & 0.1356  & 0.0571  & 12.9246  & 0.3159  & 0.1507  & 0.0667  & 11.0315  & 0.3294  & 0.2832  & 0.0895  \\
          & GIAS \cite{jeon2021gradient}  & 17.2804  & 0.2774  & 0.3346  & 0.0227  & 20.4539  & 0.1724  & 0.4913  & 0.0111  & 19.0247  & 0.1862  & 0.5740  & 0.0149  \\
          & \textbf{GIFD}  & \textbf{19.3311} & \textbf{0.1700 } & \textbf{0.4503 } & \textbf{0.0151 } & \textbf{21.9281 } & \textbf{0.1137 } & \textbf{0.5765 } & \textbf{0.0082 } & \textbf{22.8055 } & \textbf{0.1030 } & \textbf{0.6970 } & \textbf{0.0067 } \\  \midrule
    \multirow{5}[0]{*}{FFHQ*} & IG \cite{geiping2020inverting}    & 15.9020  & 0.3856  & 0.2736  & 0.0273  & 17.7422  & 0.3043  & 0.3398  & 0.0174  & 14.7029  & 0.3118  & 0.3213  & 0.0358  \\
          & GI \cite{yin2021see}    & 16.2990  & 0.3537  & 0.2917  & 0.0259  & 18.5540  & 0.2388  & 0.3808  & 0.0147  & 15.0097  & 0.3232  & 0.3201  & 0.0331  \\
          & GGL \cite{li2022auditing}   & 14.2833  & 0.2514  & 0.1982  & 0.0435  & 15.5001  & 0.2309  & 0.2513  & 0.0302  & 12.3590  & 0.2556  & 0.2322  & 0.0624  \\
          & GIAS \cite{jeon2021gradient}  & 18.4619  & 0.1912  & 0.4424  & 0.0172  & 19.6763  & 0.1615  & 0.4885  & 0.0123  & 15.3798  & 0.2250  & 0.3837  & 0.0338  \\
          & \textbf{GIFD}  & \textbf{19.8847 } & \textbf{0.1534 } & \textbf{0.4979 } & \textbf{0.0120 } & \textbf{21.3981 } & \textbf{0.1148 } & \textbf{0.5446 } & \textbf{0.0098 } & \textbf{17.4005 } & \textbf{0.1634 } & \textbf{0.4614 } & \textbf{0.0220 } \\ \bottomrule
    \end{tabular}%
    }
  \label{ood_table}%
\end{table*}%
\begin{figure*}[htbp]
	
    \begin{subfigure}{0.49\linewidth}
        \begin{minipage}[t]{0.03\linewidth}
        \rotatebox{90}{\scriptsize{\textbf{~Art Painting}}}     
        \end{minipage}%
	\begin{minipage}[t]{0.165\linewidth}
        \centering
        \includegraphics[width=1.4cm]{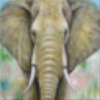}
        \end{minipage}%
        \begin{minipage}[t]{0.165\linewidth}
        \centering
        \includegraphics[width=1.4cm]{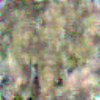}
        \end{minipage}%
        \begin{minipage}[t]{0.165\linewidth}
        \centering
        \includegraphics[width=1.4cm]{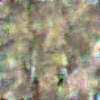}
        \end{minipage}%
        \begin{minipage}[t]{0.165\linewidth}
        \centering
        \includegraphics[width=1.4cm]{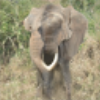}
        \end{minipage}%
        \begin{minipage}[t]{0.165\linewidth}
        \centering
        \includegraphics[width=1.4cm]{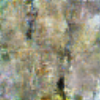}
        \end{minipage}%
        \begin{minipage}[t]{0.165\linewidth}
        \centering
        \includegraphics[width=1.4cm]{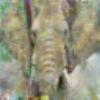}
        \end{minipage}%
        \end{subfigure}   
    \begin{subfigure}{0.49\linewidth}
        \quad
        \begin{minipage}[t]{0.165\linewidth}
        \centering
        \includegraphics[width=1.4cm]{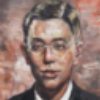}
        \end{minipage}%
        \begin{minipage}[t]{0.165\linewidth}
        \centering
        \includegraphics[width=1.4cm]{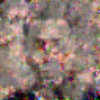}
        \end{minipage}%
        \begin{minipage}[t]{0.165\linewidth}
        \centering
        \includegraphics[width=1.4cm]{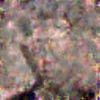}
        \end{minipage}%
        \begin{minipage}[t]{0.165\linewidth}
        \centering
        \includegraphics[width=1.4cm]{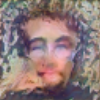}
        \end{minipage}%
        \begin{minipage}[t]{0.165\linewidth}
        \centering
        \includegraphics[width=1.4cm]{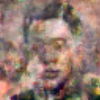}
        \end{minipage}%
        \begin{minipage}[t]{0.165\linewidth}
        \centering
        \includegraphics[width=1.4cm]{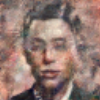}
        \end{minipage}%
        \end{subfigure}
        
    \begin{subfigure}{0.49\linewidth}
        \begin{minipage}[t]{0.03\linewidth}
        \rotatebox{90}{\scriptsize{\textbf{~~~~~~~Photo}}}         
        \end{minipage}%
        \begin{minipage}[t]{0.165\linewidth}
        \centering
        \includegraphics[width=1.4cm]{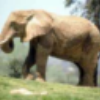}
        \end{minipage}%
        \begin{minipage}[t]{0.165\linewidth}
        \centering
        \includegraphics[width=1.4cm]{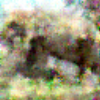}
        \end{minipage}%
        \begin{minipage}[t]{0.165\linewidth}
        \centering
        \includegraphics[width=1.4cm]{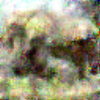}
        \end{minipage}%
        \begin{minipage}[t]{0.165\linewidth}
        \centering
        \includegraphics[width=1.4cm]{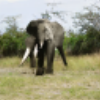}
        \end{minipage}%
        \begin{minipage}[t]{0.165\linewidth}
        \centering
        \includegraphics[width=1.4cm]{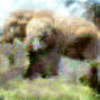}
        \end{minipage}%
        \begin{minipage}[t]{0.165\linewidth}
        \centering
        \includegraphics[width=1.4cm]{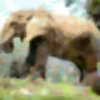}
        \end{minipage}%
        \end{subfigure}   
    \begin{subfigure}{0.49\linewidth}
        \quad
        \begin{minipage}[t]{0.165\linewidth}
        \centering
        \includegraphics[width=1.4cm]{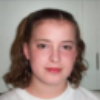}
        \end{minipage}%
        \begin{minipage}[t]{0.165\linewidth}
        \centering
        \includegraphics[width=1.4cm]{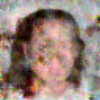}
        \end{minipage}%
        \begin{minipage}[t]{0.165\linewidth}
        \centering
        \includegraphics[width=1.4cm]{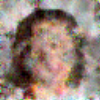}
        \end{minipage}%
        \begin{minipage}[t]{0.165\linewidth}
        \centering
        \includegraphics[width=1.4cm]{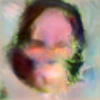}
        \end{minipage}%
        \begin{minipage}[t]{0.165\linewidth}
        \centering
        \includegraphics[width=1.4cm]{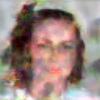}
        \end{minipage}%
        \begin{minipage}[t]{0.165\linewidth}
        \centering
        \includegraphics[width=1.4cm]{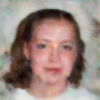}
        \end{minipage}%
        \end{subfigure}

    \begin{subfigure}{0.49\linewidth}
        \begin{minipage}[t]{0.03\linewidth}
        \rotatebox{90}{\scriptsize{\textbf{~~~Cartoon}}}     
        \end{minipage}%
        \begin{minipage}[t]{0.165\linewidth}
        \centering
        \includegraphics[width=1.4cm]{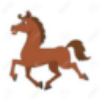}
        \centering
        \caption*{\textbf{\footnotesize{Original}}}
        \end{minipage}%
        \begin{minipage}[t]{0.165\linewidth}
        \centering
        \includegraphics[width=1.4cm]{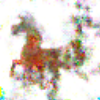}
        \centering
        \caption*{\textbf{\footnotesize{{}IG \cite{geiping2020inverting}}}}
        \end{minipage}%
        \begin{minipage}[t]{0.165\linewidth}
        \centering
        \includegraphics[width=1.4cm]{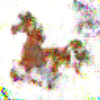}
        \centering
        \caption*{\textbf{\footnotesize{GI \cite{yin2021see}}}}
        \end{minipage}%
        \begin{minipage}[t]{0.165\linewidth}
        \centering
        \includegraphics[width=1.4cm]{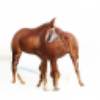}
        \centering
        \caption*{\textbf{\footnotesize{GGL \cite{li2022auditing}}}}
        \end{minipage}%
        \begin{minipage}[t]{0.165\linewidth}
        \centering
        \includegraphics[width=1.4cm]{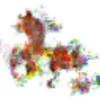}
        \centering
        \caption*{\textbf{\footnotesize{GIAS \cite{jeon2021gradient}}}}
        \end{minipage}%
        \begin{minipage}[t]{0.165\linewidth}
        \centering
        \includegraphics[width=1.4cm]{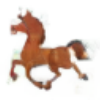}
        \centering
        \caption*{\textbf{\footnotesize{GIFD}}}
        \end{minipage}
	  \caption{ImageNet* (BigGAN)}
        \end{subfigure}
    \begin{subfigure}{0.49\linewidth}
    \quad
        \begin{minipage}[t]{0.165\linewidth}
        \centering
        \includegraphics[width=1.4cm]{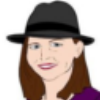}
        \centering
        \caption*{\textbf{\footnotesize{Original}}}
        \end{minipage}%
        \begin{minipage}[t]{0.165\linewidth}
        \centering
        \includegraphics[width=1.4cm]{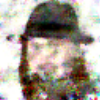}
        \centering
        \caption*{\textbf{\footnotesize{IG \cite{geiping2020inverting}}}}
        \end{minipage}%
        \begin{minipage}[t]{0.165\linewidth}
        \centering
        \includegraphics[width=1.4cm]{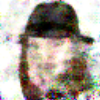}
        \centering
        \caption*{\textbf{\footnotesize{GI \cite{yin2021see}}}}
        \end{minipage}%
        \begin{minipage}[t]{0.165\linewidth}
        \centering
        \includegraphics[width=1.4cm]{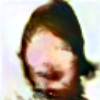}
        \centering
        \caption*{\textbf{\footnotesize{GGL \cite{li2022auditing}}}}
        \end{minipage}%
        \begin{minipage}[t]{0.165\linewidth}
        \centering
        \includegraphics[width=1.4cm]{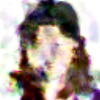}
        \centering
        \caption*{\textbf{\footnotesize{GIAS \cite{jeon2021gradient}}}}
        \end{minipage}%
        \begin{minipage}[t]{0.165\linewidth}

        \includegraphics[width=1.4cm]{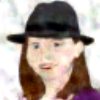}
        \centering
        \caption*{\footnotesize{\textbf{GIFD}}}
        \end{minipage}
	
        \caption{FFHQ* (StyleGAN2)}
        \end{subfigure}

    \caption{Visual comparison of different methods on ImageNet* and FFHQ*. }
    \label{ood_figure}
\end{figure*}

\noindent\textbf{Experiment Results.} By observing the results in Table \ref{main_tab}, we demonstrate that our method consistently achieves great improvement compared to the competing methods for gradient inversion attacks.  Especially in the ImageNet dataset with BigGAN, our method has nearly 2.5dB and 0.1 improvements in average PSNR and LPIPS values respectively. As the visualization comparison shown in Figure \ref{main_figure}, under a more practical setting, most existing methods struggle to recover meaningful and high-quality images even at $B=1$, while our method reveals significant information about the private data and achieves pixel-level reconstruction on both two datasets. 

The GAN-based methods (i.e. GGL, GIAS, GIFD) generally achieve better results than the GAN-free methods (i.e. GI, IG) on the FFHQ dataset. This indicates that the special data distribution of human-face can be more easily learned by the generative model so that the gain from the GAN prior is larger. We also observe that the GAN-based method GGL, which only optimizes the latent code and does not fully exploit the GAN prior, yields unsatisfactory results and performs even worse than the GAN-free methods \cite{geiping2020inverting, yin2021see} on the ImageNet dataset, which again verifies the necessity of searching intermediate layers. 

We note that the performance of GIAS with BigGAN is worse than with StyleGAN2. One reason is that the data of ImageNet is more diverse. More importantly, with such a large number of parameters in BigGAN, the solution space for the GAN parameter search process becomes larger and presents a great challenge, \ie, GIAS is more susceptible to the scale of GAN. In contrast, GIFD chooses to optimize the intermediate features and then avoids this problem, hence achieving faithful reconstruction on both two GANs, demonstrating the excellent versatility of our method.



\subsection{Out of Distribution Data Recovery}
We then consider a more practical scenario where the training sets of the GAN model and the FL task obey different data distributions. Considering the difficulty and feasibility of gradient attack tasks, we define the OOD data as having the same label space, but quite different feature distributions. Hereinafter, we denote the OOD data of ImageNet and FFHQ by ImageNet* and FFHQ* respectively.

PAC \cite{li2017deeper} dataset is a widely used benchmark for domain generalization with four different styles, \ie, Art Painting, Cartoon, Photo, and Sketch. In order to achieve our OOD setting, we manually select data with three different styles (\ie, Art Painting, Cartoon, Photo) from the validation set of PACS. For each style in ImageNet*, we select 15 images of guitar, elephant and horse in total. For FFHQ*, we select 15 images for each style and crop them to obtain the face images. We present visual comparison and quantitative results in Figure \ref{ood_figure} and Table \ref{ood_table}.

\noindent\textbf{Experiment Results.} As shown in Table \ref{ood_table}, the experiment results demonstrate our significant improvement over the baseline methods. For instance, our method has nearly 3.8dB improvement in average PSNR upon GIAS for Cartoon in ImageNet*. Compared with other styles, the GAN-based methods perform best on Photo, whose domain characteristics are similar to the training sets of GANs. We also note that for Art in ImageNet*, the GAN-based methods except GIFD perform even worse than the GAN-free ones, which implies that here the gain from GAN is minor and even brings negative effects to them.

Generally, the other GAN-based methods preserve more pre-trained knowledge from ImageNet or FFHQ, thus struggling to generate images similar to ground truth with different styles. In contrast, our method augments the generative ability of the GAN models and enlarges the diversity of the output space, hence achieving outstanding performance. Thus, with our proposed GIFD, we are able to safely relax the assumption that the datasets of the generative model and FL have to obey the same feature distribution.

\subsection{Attacks under Certain Defense Strategies}

Next, we consider attacking a more robust and secure FL system with defense strategies. In order to make a fair comparison, we equip all the baselines with the well-designed gradient transformation technique mentioned before to mitigate the impact of defense.

We consider a relatively strict defense setup as the previous work \cite{li2017deeper}: (1) \textit{Gaussian Noise} with standard deviation 0.1; (2) \textit{Gradient Clipping} with a clip bound of 4; (3) \textit{Gradient Sparsification} in a sparsity of 90; and (4) \textit{Soteria} with a pruning rate of 80\%.
\begin{table}[htbp]

  \caption{PSNR mean of different methods under different defense strategies. }
    \begin{subtable}[t]{\linewidth}
    \resizebox{\linewidth}{!}{\begin{tabular}{lcccc}
    \toprule
    \multicolumn{1}{l}{\multirow{2}[0]{*}{Method}} & \multicolumn{4}{c}{Defense Strategies} \\ \cmidrule(lr){2-5}
    & Noise \cite{geyer2017differentially} & Clipping \cite{geyer2017differentially} & Sparsification \cite{aji-heafield-2017-sparse} & Soteria \cite{sun2021soteria} \\ \midrule
       IG \cite{geiping2020inverting}   & 11.0654  & 16.4418  & 12.0760  & 9.1941  \\
           GI \cite{yin2021see}    & 10.0818  & 12.5387  & 12.1691  & 10.1831\\
           GGL \cite{li2022auditing}   & 12.7640  & 12.7930  & 12.6810  & 12.8433  \\
           GIAS \cite{jeon2021gradient}  & 12.5397  & 17.9384  & 15.1745  & 16.8151  \\
           GIFD  & \textbf{13.2558} & \textbf{18.8983} & \textbf{16.0240} & \textbf{18.3205} \\ \bottomrule
    \end{tabular}
    }
    \caption{ImageNet}
    \end{subtable}
    
    \begin{subtable}[t]{\linewidth}
    \resizebox{\linewidth}{!}{\begin{tabular}{lcccc} \toprule
    \multicolumn{1}{l}{\multirow{2}[0]{*}{Method}} & \multicolumn{4}{c}{Defense Strategies} \\ \cmidrule(lr){2-5}
    & Noise \cite{geyer2017differentially} & Clipping \cite{geyer2017differentially} & Sparsification \cite{aji-heafield-2017-sparse} & Soteria \cite{sun2021soteria} \\ \midrule
        IG \cite{geiping2020inverting}    & 11.2766   & 18.1382  & 12.0077  & 9.8334 \\
           GI \cite{yin2021see}    & 10.4968  & 12.4146  & 12.1849  & 10.0843  \\
           GGL \cite{li2022auditing}   & \textbf{14.8982}   & 15.6669  & 14.9123  & 15.1798  \\
           GIAS \cite{jeon2021gradient}  & 12.1276   & 20.4726  & 16.7005  & 20.4283  \\
           GIFD  & 13.7118 
 & \textbf{21.2861} & \textbf{17.3253} & \textbf{21.1545} \\ \bottomrule
    \end{tabular}
    }
    \caption{FFHQ}
    \end{subtable}
    
  \label{defense_table}%
\end{table}%

\noindent\textbf{Experiment Results.} We present experiment results in Table \ref{defense_table} compared to related methods. In general, with the underlying gradient transformation and the fully exploited GAN image prior, GIFD is still able to invert a degraded gradient observation to generate high-quality images or reveal private information, especially in cases of clipping and Soteria. One exception is that GGL takes the lead on FFHQ when applying additive noise operation. This is because the gradient information is seriously corrupted by the added high-variance Gaussian noise and is no more enough for pixel-level reconstruction. However, GGL only searches the latent space and with GAN's powerful generative capability, it can still produce well-formed images with clear facial contour, which can give a fair result in the metrics even though they are quite different from the original ones. This also indicates that adding Gaussian noise is indeed an effective defense method against related attacks when the variance exceeds a certain threshold.

\subsection{Performance of Larger Batch Sizes}

We then increase the batch size and observe the results of each algorithm. Notably, we assume that no duplicate labels in each batch and infer the labels from the received gradients \cite{yin2021see}. We present the results on ImageNet in Table \ref{bs}, see Appendix for results on FFHQ.

\begin{table}[htbp]
  \centering
  \caption{PSNR mean of different methods for different batch sizes on ImageNet.}
    \resizebox{\linewidth}{!}{\begin{tabular}{lrrrrrr} \toprule
    \multicolumn{1}{l}{\multirow{2}[0]{*}{Method}} & \multicolumn{6}{c}{Batch Size} \\ \cmidrule(lr){2-7}
          & \multicolumn{1}{c}{1} & \multicolumn{1}{c}{2} & \multicolumn{1}{c}{4} & \multicolumn{1}{c}{8} & \multicolumn{1}{c}{16} & \multicolumn{1}{c}{32} \\ \midrule
    IG\cite{geiping2020inverting}    & 17.4634  & 15.2417  & 14.3744  & 13.6599  & 13.1545  & 12.0795  \\
    GI\cite{yin2021see}    & 17.4373  & 14.7293  & 14.0947  & 13.3001  & 12.7842  & 11.8767  \\
    GGL\cite{li2022auditing}   & 12.7511  & 12.8903  & 13.1875  & 12.6001  & 11.8027  & 11.0896  \\
    GIAS\cite{jeon2021gradient}  & 17.1401  & 16.1683  & 15.5894  & 15.2130  & 14.4462  & 13.6080  \\
    \textbf{GIFD}  & \textbf{20.6217}  & \textbf{16.7542} & \textbf{16.4272}  & \textbf{15.4889}  & \textbf{14.6500}  & \textbf{13.8106}  \\
    \bottomrule
    \end{tabular}%
    
    }
  \label{bs}%
\end{table}%

\noindent\textbf{Experiment Results.} We find that the proposed GIFD achieves a steady improvement over previous methods at any batch size. The numerical results also show that the performance of all methods generally degrades as the batch size increases, implying that the reconstruction at large batch sizes is still a significant challenge.

\section{Conclusion}
We propose GIFD, a powerful gradient inversion attack that can generalize well in unseen OOD data scenarios. We leverage the GAN prior via optimizing the feature domain of the generative model to generate stable and high-fidelity inversion results. Through extensive experiments, we demonstrate the effectiveness of GIFD with two widely used pre-trained GANs on two large datasets in a variety of more practical and challenging scenarios. To alleviate the proposed threat, one possible defense strategy is utilizing gradient-based adversarial noise as a novel privacy mechanism to provide confused inversion. 

We hope this paper can inspire some new ideas for future work and make contributions to the gradient attacks under more realistic scenarios. We also hope that our work can shed light on the design of privacy mechanisms, to enhance the security and robustness of FL systems.

{\small
\bibliographystyle{ieee_fullname}
\bibliography{egbib}
}

\clearpage
\newpage

\section*{A. Larger Batch Sizes on FFHQ}
\label{largerB}
We provide the results of different batch sizes on the FFHQ dataset. Since the label extraction algorithm\cite{yin2021see} requires non-repeating labels in a batch, the batch size cannot exceed the number of categories. Therefore, the maximum batch size of our experiment is $8$ on FFHQ. Note that the latent vector of StyleGAN2 has a relatively large number of parameters to be optimized and the CMA-ES optimizer, adopted by GGL, does not support large-scale optimization. Thus, GGL is unable to operate when $B>2$.

\begin{table}[htbp]
  \centering
  \caption{PSNR mean of different methods for different batch sizes on FFHQ.}
    \resizebox{0.85\linewidth}{!}{\begin{tabular}{lcccc} \toprule
    \multicolumn{1}{l}{\multirow{2}[0]{*}{Method}} & \multicolumn{4}{c}{Batch Size} \\ \cmidrule(lr){2-5}
          & \multicolumn{1}{c}{1} & \multicolumn{1}{c}{2} & \multicolumn{1}{c}{4} & \multicolumn{1}{c}{8} \\ \midrule
    IG \cite{geiping2020inverting}    & 19.07606 & 16.27659 & 13.87481 & 12.24488 \\
    GI \cite{yin2021see}    & 17.35061 & 15.55461 & 13.42360 & 12.19926 \\
    GGL \cite{li2022auditing}   & 14.74791 & 13.35473 &   ——   &  ——\\
    GIAS \cite{jeon2021gradient}  & 20.07786 & 16.95568 & 13.67158 & 12.49889 \\
    GIFD  & \textbf{21.13338} & \textbf{17.96191} & \textbf{14.34927} & \textbf{12.74023} \\ \bottomrule
    \end{tabular}%
    }

  \label{ffhq_bs}%
\end{table}%

As shown in Table \ref{ffhq_bs}, GIFD outperforms all previous methods at every batch size we considered. During the label extraction process, the error rate of inferred labels is relatively high when $B > 2$, leading to degraded performance of all methods because of losing the significant information brought by the correct labels.

\section*{B. Inference Speed Comparison}
\begin{figure}[htbp]
	\centering
	\includegraphics[width=0.8\linewidth]{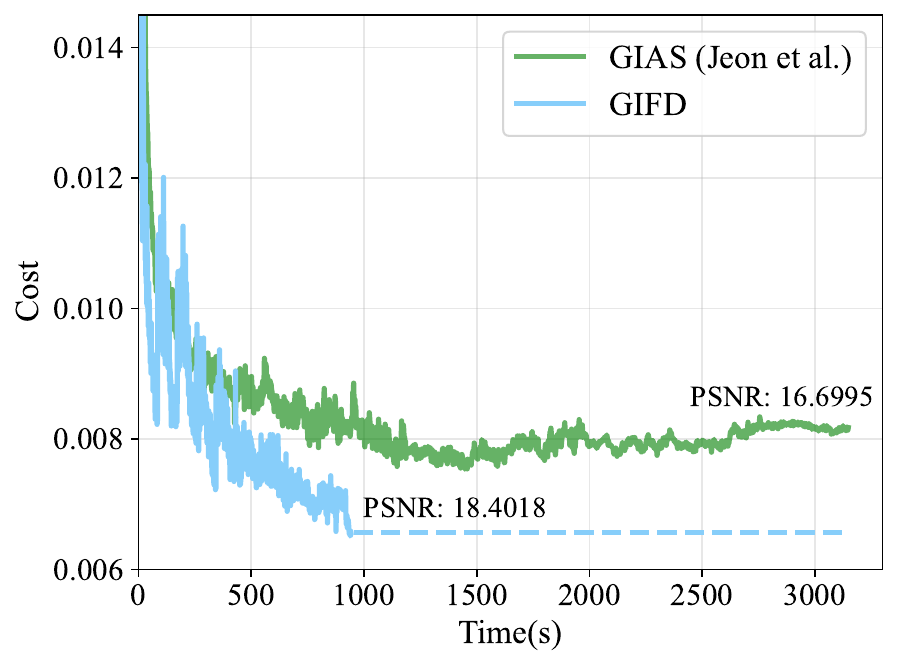}

	\centering
	\caption{The cost function over time of GIAS and GIFD with $B=4$. We give 4 trials and calculate the average values. For a fair comparison, both methods execute a total of 8000 iterations.}
	\label{cost-function curve}
\end{figure}

GIAS \cite{jeon2021gradient}, searching the latent and parameter space of the generative model in turn, generally performs best among the previous methods. A series of experiments have demonstrated that our method achieves consistent improvement over GIAS. Besides, GIFD only searches the feature domain, whose optimized parameters are far less compared with the generator's parameters. And GIAS requires a specific generator to be trained for each reconstructed image, consuming great inference time and GPU memory. Thus GIFD should have an advantage in inference speed. In Figure \ref{cost-function curve}, we draw the cost-time curve for the intermediate feature searching phase of GIFD and the parameter space searching phase of GIAS. The corresponding PSNR values of the results are also annotated in the figure. 

As expected, GIFD completes the optimization task only with less than $1/3$ of the time for GIAS and further reduces the loss. We also find that every time the optimizer moves to the next feature space, the loss comes to a peak and quickly decreases over several iterations. After getting stable, the loss value of each layer is always below the curve of GIAS. 

Further, we conduct experiments on methods with the same distance metric as GIFD in Table \ref{timecost}. Although IG converges faster, an attacker can perform attacks offline with a copy of the historical global model and observed gradients, thus the attack effect is more essential than inference speed. In this case, GIFD achieves a good trade-off between effectiveness and time cost.

\begin{table}[h]
  \centering
 \scriptsize 
  \caption{Converge time and results at batch size 4.}
    \resizebox{6.5cm}{!}{\begin{tabular}{lccc} \toprule
    Method & IG \cite{geiping2020inverting}    & GIAS \cite{jeon2021gradient}  & GIFD \\     \midrule
    Time(s)↓ & \textbf{726.8008} & 1360.7808 & 907.727 \\
    PSNR↑  & 14.1896 & 16.6995 & \textbf{18.4018} \\
    Loss↓  & 0.008029 & 0.007803 & \textbf{0.006865} \\ \bottomrule
    \end{tabular}%
    }
  \label{timecost}%
\end{table}%

\section*{C. More FL Global Models.} To further validate our method, We provide numerical results of PSNR on more global models below.
\begin{table}[htbp]
  \centering
\captionsetup{}
  \caption{PSNR for different global models on ImageNet.}
    \resizebox{8.5cm}{!}{\begin{tabular}{lccccc} \toprule
    Global model & \multicolumn{1}{c}{IG \cite{geiping2020inverting}} & \multicolumn{1}{c}{GI \cite{yin2021see}} & \multicolumn{1}{c}{GGL \cite{li2022auditing}} & \multicolumn{1}{c}{GIAS \cite{jeon2021gradient}} & \multicolumn{1}{c}{\textbf{GIFD}} \\ \midrule
    ConvNet & 22.8043 &  21.3876     & 13.2582      &  24.1741  & \textbf{25.5646} \\
    AlexNet & 15.3390      &  16.3423     & 13.6605      &  17.5883     & \textbf{19.7926} \\
    VGG-16 & 13.3505     &  13.3856     &  13.9808     &  14.5414   & \textbf{16.0014} \\
    ResNet-18 & 17.0756      & 16.5109  & 13.3885    & 17.4923      & \textbf{20.0534}       \\
    DenseNet-121 &  18.0840    & 17.3253      & 14.3538      &  18.1686     & \textbf{19.7376} \\ \bottomrule
    \end{tabular}%
    }
  \label{globalmodel}%
\end{table}%

As shown in Table \ref{globalmodel}, the overall attack performance varied across different global models and GIFD always performs the best, convincing the superiority of our method. It also implies that the model structure is related to the defense effect. Further study can look into it and design more secure model structures.

\section*{D. More Visual Comparison}
\begin{figure}[htbp]
    \begin{subfigure}{1\linewidth}
        \quad
        \begin{minipage}[t]{0.165\linewidth}
        \centering
        \includegraphics[width=1.35cm]{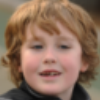}
        \end{minipage}%
        \begin{minipage}[t]{0.165\linewidth}
        \centering
        \includegraphics[width=1.35cm]{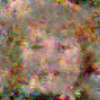}
        \end{minipage}%
        \begin{minipage}[t]{0.165\linewidth}
        \centering
        \includegraphics[width=1.35cm]{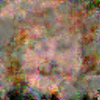}
        \end{minipage}%
        \begin{minipage}[t]{0.165\linewidth}
        \centering
        \includegraphics[width=1.35cm]{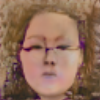}
        \end{minipage}%
        \begin{minipage}[t]{0.165\linewidth}
        \centering
        \includegraphics[width=1.35cm]{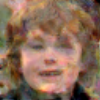}
        \end{minipage}%
        \begin{minipage}[t]{0.165\linewidth}
        \centering
        \includegraphics[width=1.35cm]{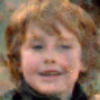}
        \end{minipage}%
        \end{subfigure}
    \begin{subfigure}{1\linewidth}
    \quad
        \begin{minipage}[t]{0.165\linewidth}
        \centering
        \includegraphics[width=1.35cm]{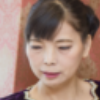}
        \centering
        \end{minipage}%
        \begin{minipage}[t]{0.165\linewidth}
        \centering
        \includegraphics[width=1.35cm]{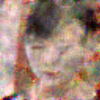}
        \centering
        \end{minipage}%
        \begin{minipage}[t]{0.165\linewidth}
        \centering
        \includegraphics[width=1.35cm]{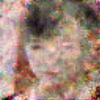}
        \centering
        \end{minipage}%
        \begin{minipage}[t]{0.165\linewidth}
        \centering
        \includegraphics[width=1.35cm]{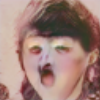}
        \centering
        \end{minipage}%
        \begin{minipage}[t]{0.165\linewidth}
        \centering
        \includegraphics[width=1.35cm]{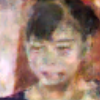}
                \centering
        \end{minipage}%
        \begin{minipage}[t]{0.165\linewidth}

        \includegraphics[width=1.35cm]{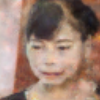}
        \centering
        \end{minipage}
	
        \end{subfigure}
    \begin{subfigure}{1\linewidth}
    \quad
        \begin{minipage}[t]{0.165\linewidth}
        \centering
        \includegraphics[width=1.35cm]{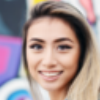}
        \centering
        \end{minipage}%
        \begin{minipage}[t]{0.165\linewidth}
        \centering
        \includegraphics[width=1.35cm]{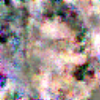}
        \centering
        \end{minipage}%
        \begin{minipage}[t]{0.165\linewidth}
        \centering
        \includegraphics[width=1.35cm]{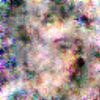}
        \centering
        \end{minipage}%
        \begin{minipage}[t]{0.165\linewidth}
        \centering
        \includegraphics[width=1.35cm]{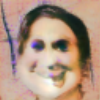}
        \centering
        \end{minipage}%
        \begin{minipage}[t]{0.165\linewidth}
        \centering
        \includegraphics[width=1.35cm]{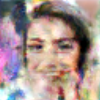}
                \centering
        \end{minipage}%
        \begin{minipage}[t]{0.165\linewidth}

        \includegraphics[width=1.35cm]{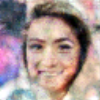}
        \centering
        \end{minipage}
	
        \end{subfigure}
    \begin{subfigure}{1\linewidth}
        \quad
        \begin{minipage}[t]{0.165\linewidth}
        \centering
        \includegraphics[width=1.35cm]{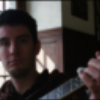}
        \end{minipage}%
        \begin{minipage}[t]{0.165\linewidth}
        \centering
        \includegraphics[width=1.35cm]{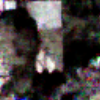}
        \end{minipage}%
        \begin{minipage}[t]{0.165\linewidth}
        \centering
        \includegraphics[width=1.35cm]{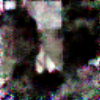}
        \end{minipage}%
        \begin{minipage}[t]{0.165\linewidth}
        \centering
        \includegraphics[width=1.35cm]{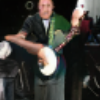}
        \end{minipage}%
        \begin{minipage}[t]{0.165\linewidth}
        \centering
        \includegraphics[width=1.35cm]{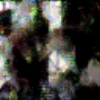}
        \end{minipage}%
        \begin{minipage}[t]{0.165\linewidth}
        \centering
        \includegraphics[width=1.35cm]{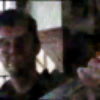}
        \end{minipage}%
        \end{subfigure}  
    \begin{subfigure}{1\linewidth}
    \quad
        \begin{minipage}[t]{0.165\linewidth}
        \centering
        \includegraphics[width=1.35cm]{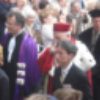}
        \centering
        \end{minipage}%
        \begin{minipage}[t]{0.165\linewidth}
        \centering
        \includegraphics[width=1.35cm]{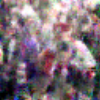}
        \centering
        \end{minipage}%
        \begin{minipage}[t]{0.165\linewidth}
        \centering
        \includegraphics[width=1.35cm]{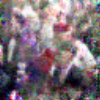}
        \centering
        \end{minipage}%
        \begin{minipage}[t]{0.165\linewidth}
        \centering
        \includegraphics[width=1.35cm]{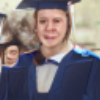}
        \centering
        \end{minipage}%
        \begin{minipage}[t]{0.165\linewidth}
        \centering
        \includegraphics[width=1.35cm]{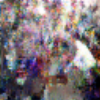}
                \centering
        \end{minipage}%
        \begin{minipage}[t]{0.165\linewidth}

        \includegraphics[width=1.35cm]{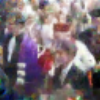}
        \centering
        \end{minipage}
	
        \end{subfigure}
    \begin{subfigure}{1\linewidth}
    \quad
        \begin{minipage}[t]{0.165\linewidth}
        \centering
        \includegraphics[width=1.35cm]{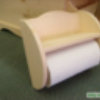}
        \centering
        \caption*{\textbf{Original}}
        \end{minipage}%
        \begin{minipage}[t]{0.165\linewidth}
        \centering
        \includegraphics[width=1.35cm]{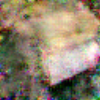}
        \centering
        \caption*{\textbf{IG \cite{geiping2020inverting}}}
        \end{minipage}%
        \begin{minipage}[t]{0.165\linewidth}
        \centering
        \includegraphics[width=1.35cm]{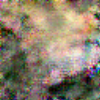}
        \centering
        \caption*{\textbf{GI \cite{yin2021see}}}
        \end{minipage}%
        \begin{minipage}[t]{0.165\linewidth}
        \centering
        \includegraphics[width=1.35cm]{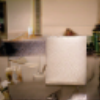}
        \centering
        \caption*{\textbf{GGL \cite{li2022auditing}}}
        \end{minipage}%
        \begin{minipage}[t]{0.165\linewidth}
        \centering
        \includegraphics[width=1.35cm]{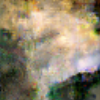}
                \centering
        \caption*{\textbf{GIAS \cite{jeon2021gradient}}}
        \end{minipage}%
        \begin{minipage}[t]{0.165\linewidth}

        \includegraphics[width=1.35cm]{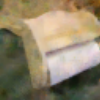}
        \centering
        \caption*{\textbf{GIFD}}
        \end{minipage}
	
        \caption{In-distribution Data}
        \end{subfigure}

    \begin{subfigure}{1\linewidth}
        \begin{minipage}[t]{0.03\linewidth}
        \rotatebox{90}{\scriptsize{\textbf{Art Painting}}}         
        \end{minipage}%
        \begin{minipage}[t]{0.165\linewidth}
        \centering
        \includegraphics[width=1.35cm]{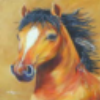}
        \end{minipage}%
        \begin{minipage}[t]{0.165\linewidth}
        \centering
        \includegraphics[width=1.35cm]{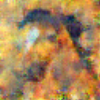}
        \end{minipage}%
        \begin{minipage}[t]{0.165\linewidth}
        \centering
        \includegraphics[width=1.35cm]{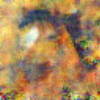}
        \end{minipage}%
        \begin{minipage}[t]{0.165\linewidth}
        \centering
        \includegraphics[width=1.35cm]{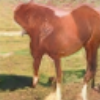}
        \end{minipage}%
        \begin{minipage}[t]{0.165\linewidth}
        \centering
        \includegraphics[width=1.35cm]{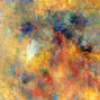}
        \end{minipage}%
        \begin{minipage}[t]{0.165\linewidth}
        \centering
        \includegraphics[width=1.35cm]{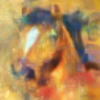}
        \end{minipage}%
        \end{subfigure} 
    \begin{subfigure}{1\linewidth}
        \begin{minipage}[t]{0.03\linewidth}
        \rotatebox{90}{\scriptsize{}}     
        \end{minipage}%
	\begin{minipage}[t]{0.165\linewidth}
        \centering
        \includegraphics[width=1.35cm]{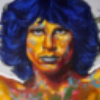}
        \end{minipage}%
        \begin{minipage}[t]{0.165\linewidth}
        \centering
        \includegraphics[width=1.35cm]{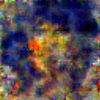}
        \end{minipage}%
        \begin{minipage}[t]{0.165\linewidth}
        \centering
        \includegraphics[width=1.35cm]{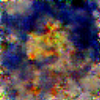}
        \end{minipage}%
        \begin{minipage}[t]{0.165\linewidth}
        \centering
        \includegraphics[width=1.35cm]{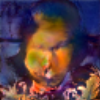}
        \end{minipage}%
        \begin{minipage}[t]{0.165\linewidth}
        \centering
        \includegraphics[width=1.35cm]{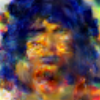}
        \end{minipage}%
        \begin{minipage}[t]{0.165\linewidth}
        \centering
        \includegraphics[width=1.35cm]{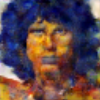}
        \end{minipage}%
        \end{subfigure}
    \begin{subfigure}{1\linewidth}
        \begin{minipage}[t]{0.03\linewidth}
        \rotatebox{90}{\scriptsize{\textbf{~~~~~~~Photo}}}         
        \end{minipage}%
        \begin{minipage}[t]{0.165\linewidth}
        \centering
        \includegraphics[width=1.35cm]{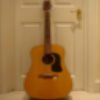}
        \end{minipage}%
        \begin{minipage}[t]{0.165\linewidth}
        \centering
        \includegraphics[width=1.35cm]{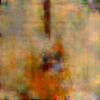}
        \end{minipage}%
        \begin{minipage}[t]{0.165\linewidth}
        \centering
        \includegraphics[width=1.35cm]{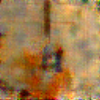}
        \end{minipage}%
        \begin{minipage}[t]{0.165\linewidth}
        \centering
        \includegraphics[width=1.35cm]{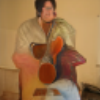}
        \end{minipage}%
        \begin{minipage}[t]{0.165\linewidth}
        \centering
        \includegraphics[width=1.35cm]{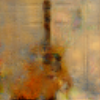}
        \end{minipage}%
        \begin{minipage}[t]{0.165\linewidth}
        \centering
        \includegraphics[width=1.35cm]{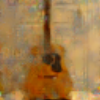}
        \end{minipage}%
        \end{subfigure} 
    \begin{subfigure}{1\linewidth}
        \begin{minipage}[t]{0.03\linewidth}
        \rotatebox{90}{\scriptsize{\textbf{}}}     
        \end{minipage}%
	\begin{minipage}[t]{0.165\linewidth}
        \centering
        \includegraphics[width=1.35cm]{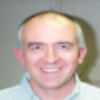}
        \end{minipage}%
        \begin{minipage}[t]{0.165\linewidth}
        \centering
        \includegraphics[width=1.35cm]{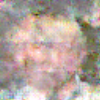}
        \end{minipage}%
        \begin{minipage}[t]{0.165\linewidth}
        \centering
        \includegraphics[width=1.35cm]{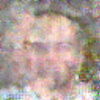}
        \end{minipage}%
        \begin{minipage}[t]{0.165\linewidth}
        \centering
        \includegraphics[width=1.35cm]{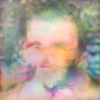}
        \end{minipage}%
        \begin{minipage}[t]{0.165\linewidth}
        \centering
        \includegraphics[width=1.35cm]{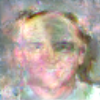}
        \end{minipage}%
        \begin{minipage}[t]{0.165\linewidth}
        \centering
        \includegraphics[width=1.35cm]{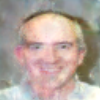}
        \end{minipage}%
        \end{subfigure}

    \begin{subfigure}{1\linewidth}
        \begin{minipage}[t]{0.03\linewidth}
        \rotatebox{90}{\scriptsize{\textbf{~~~Cartoon}}}         
        \end{minipage}%
        \begin{minipage}[t]{0.165\linewidth}
        \centering
        \includegraphics[width=1.35cm]{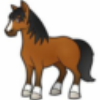}
        \end{minipage}%
        \begin{minipage}[t]{0.165\linewidth}
        \centering
        \includegraphics[width=1.35cm]{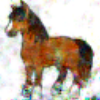}
        \end{minipage}%
        \begin{minipage}[t]{0.165\linewidth}
        \centering
        \includegraphics[width=1.35cm]{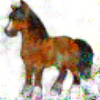}
        \end{minipage}%
        \begin{minipage}[t]{0.165\linewidth}
        \centering
        \includegraphics[width=1.35cm]{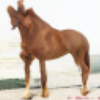}
        \end{minipage}%
        \begin{minipage}[t]{0.165\linewidth}
        \centering
        \includegraphics[width=1.35cm]{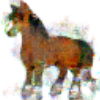}
        \end{minipage}%
        \begin{minipage}[t]{0.165\linewidth}
        \centering
        \includegraphics[width=1.35cm]{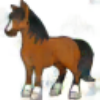}
        \end{minipage}%
        \end{subfigure}
    \begin{subfigure}{1\linewidth}
        \begin{minipage}[t]{0.03\linewidth}
        \rotatebox{90}{\scriptsize{\textbf{}}}     
        \end{minipage}%
        \begin{minipage}[t]{0.165\linewidth}
        \centering
        \includegraphics[width=1.35cm]{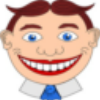}
        \centering
        \caption*{\textbf{Original}}
        \end{minipage}%
        \begin{minipage}[t]{0.165\linewidth}
        \centering
        \includegraphics[width=1.35cm]{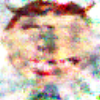}
        \centering
        \caption*{\textbf{IG \cite{geiping2020inverting}}}
        \end{minipage}%
        \begin{minipage}[t]{0.165\linewidth}
        \centering
        \includegraphics[width=1.35cm]{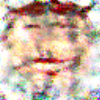}
        \centering
        \caption*{\textbf{GI \cite{yin2021see}}}
        \end{minipage}%
        \begin{minipage}[t]{0.165\linewidth}
        \centering
        \includegraphics[width=1.35cm]{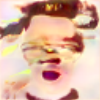}
        \centering
        \caption*{\textbf{GGL \cite{li2022auditing}}}
        \end{minipage}%
        \begin{minipage}[t]{0.165\linewidth}
        \centering
        \includegraphics[width=1.35cm]{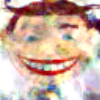}
        \centering
        \caption*{\textbf{GIAS \cite{jeon2021gradient}}}
        \end{minipage}%
        \begin{minipage}[t]{0.165\linewidth}
        \centering
        \includegraphics[width=1.35cm]{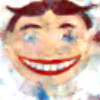}
        \centering
        \caption*{\textbf{GIFD}}
        \end{minipage}
	  \caption{Out-of-distribution Data}
        \end{subfigure}

    \caption{More visual comparison of different methods on in-distribution and out-of-distribution data.}
    \label{more}
\end{figure}

We show more qualitative comparison on both in-distribution datasets (ImageNet\cite{deng2009imagenet} and FFHQ\cite{karras2019style}) and out-of-distribution data (PACS \cite{li2017deeper}) in Figure \ref{more}. 

We can easily find that our reconstructed images stay closer to the original manifold. This again provides evidence that our method fully exploits the generative model as an image prior and hence could reveal more sensitive information about the private data.

\section*{E. Ablation Study}

We conduct ablation experiments on both two datasets to further verify the effectiveness of each proposed technique. There are three variants of GIFD. GIFD-$z$ only searches the latent space. GIFD-$f$ starts to search the intermediate feature domain without the $l_1$ ball limitation and outputs the final results from the last searched intermediate layer. Based on GIFD-$f$, GIFD-$e$ selects outputs from the layer with the least matching error. And GIFD is GIFD-$e$ plus the $l_1$ ball limitation. The results of Table \ref{ablation_table} show that each aforementioned technique can further improve the performance.

\begin{table}[htbp]
  \centering
  \caption{Ablation study of GIFD and its three variants on every 1000th image of ImageNet and FFHQ validation set.}
  
  \centering
    \begin{subtable}[t]{\linewidth}
    \centering
     \resizebox{0.8\linewidth}{!}{
     \begin{tabular}{lccccc} \toprule
    \multicolumn{2}{l}{\multirow{2}[1]{*}{Method}} & \multicolumn{4}{c}{Metric} \\ \cmidrule(lr){3-6}
     & & \multicolumn{1}{c}{PSNR$\uparrow$} & \multicolumn{1}{c}{LPIPS$\downarrow$} & \multicolumn{1}{c}{SSIM$\uparrow$} & \multicolumn{1}{c}{MSE$\downarrow$} \\  \midrule
    \multicolumn{2}{l}{GIFD-$z$} & 13.9451  & 0.3445  & 0.1463  & 0.0488  \\
    \multicolumn{2}{l}{GIFD-$f$} & 18.6457  & 0.2320  & 0.3916  & 0.0180  \\
    \multicolumn{2}{l}{GIFD-$e$} & 19.4662  & 0.1900  & 0.4383  & 0.0161  \\
    \multicolumn{2}{l}{GIFD} & \textbf{20.0534} & \textbf{0.1559} & \textbf{0.4713} & \textbf{0.0141} \\ \bottomrule
    
    \end{tabular}%
    }
    \caption{ImageNet}
    \end{subtable}

    \centering
    \begin{subtable}[t]{\linewidth}
    \centering
     \resizebox{0.8\linewidth}{!}{
     \begin{tabular}{lccccc} \toprule
    \multicolumn{2}{l}{\multirow{2}[1]{*}{Method}} & \multicolumn{4}{c}{Metric} \\ \cmidrule(lr){3-6}
     & & \multicolumn{1}{c}{PSNR$\uparrow$} & \multicolumn{1}{c}{LPIPS$\downarrow$} & \multicolumn{1}{c}{SSIM$\uparrow$} & \multicolumn{1}{c}{MSE$\downarrow$} \\  \midrule
    \multicolumn{2}{l}{GIFD-$z$} & 16.9947  & 0.1351  & 0.3931  & 0.0263  \\
    \multicolumn{2}{l}{GIFD-$f$} & 20.2506  & 0.1462  & 0.5210  & 0.0123  \\
    \multicolumn{2}{l}{GIFD-$e$} & 20.5839  & 0.1267  & 0.5412  & 0.0119  \\
    \multicolumn{2}{l}{GIFD} &  \textbf{21.3368} & \textbf{0.1023} & \textbf{0.5768} & \textbf{0.0098} \\ \bottomrule
    
    \end{tabular}%
    }
    \caption{FFHQ}
    \end{subtable}

\label{ablation_table}%
\end{table}%

\section*{F. Another Approach for OOD Problem}
Motivated by previous work \cite{hitaj2017deep, wang2019beyond}, Jeon \etal \cite{jeon2021gradient} propose another method that can solve the out-of-distribution (OOD) problem through training a generative model only with the shared gradients, \ie, Gradient Inversion to Meta-Learn (GIML). They regard the global model in FL framework as a discriminator and train the generator by solving a set of gradient inversion tasks. However, assuming that the private labels are known, their experiments are limited to $32\times32$ images and the improvement is also limited. On the contrary, our method has impressive effects on OOD data and only requires a pre-trained GAN, which releases the computing costs of training a generative model.

\section*{G. Details about Gradient Transformation}
More specifically, the adversary can infer three defense strategies as follows (denote the received gradients by $g$):

(1) \textit{Gradient clipping.} Given a clipping bound $c$, gradient clipping transforms the gradients as $\mathcal{T}(g,c) = g\cdot\min(\frac{c}{\Vert g\Vert_2},1)$. Since this operation is always layer-wise, the attacker can compute the $\ell_2$ norm at each layer of the received gradients as the estimated clipping bound.

(2) \textit{Gradient sparsification.} Given a pruning rate $p\in(0,1)$, the client only transmits the $(1 - p)$ largest values of $g$ (in absolute value) and the rest values are replaced by zero. Implemented by applying a layer-wise mask, the gradient sparsity can be estimated by observing the percentage of non-zero entries in the shared gradients.

(3) \textit{Soteria.} Recently proposed by \cite{sun2021soteria}, it is an efficient and reliable defense strategy. This operation is actually equivalent to applying a mask only to the gradients of the defended layer. Once the global model $f_\theta$ and input $\mathbf{x}$ are given, this process becomes deterministic. Then, the attacker can inverse this mask according to the non-zero entries of the gradients from the defended layer.

\section*{H. Discussion}
\textbf{Reconstruction at large batch sizes.} Although our method performs well in a range of experiments, the improvement at large batch sizes is still limited, implying that attacks in such a scenario are still a major challenge. Meanwhile, we make the assumption that there are no duplicate labels in each batch to infer the labels, which is also difficult to achieve in the real scenario. To relax the assumption that non-repeated labels in a batch, we notice a recent paper \cite{ma2022instance} has addressed the problem effectively, which can be perfectly combined with our GIFD to enhance the attack when there are duplicate labels.

\textbf{Hypothesis about OOD data.} In order to utilize the powerful information brought by labels, we assume that images from different distributions have the same label space. To tackle more realistic OOD problems where label spaces are different, subsequent research could consider using more powerful diffusion models \cite{ho2020denoising, nichol2021improved} as prior information, or using other related techniques to improve the expressiveness of generative models.

\section*{I. Experimental Details}
\label{exper}
For each intermediate feature domain, we use Adam optimizer with $0.1$ as the initial learning rate and give $1000$ iterations. We adopt the warm-up strategy, where the learning rate linearly warms up from $0$ to $0.1$ during the first $1/20$ of the optimization and gradually decays to $0$ in the last $3/4$ stage using cosine decay.

Guided by the theory \cite{daras2021intermediate} that a sequence of increasing radii of the $l_1$ ball tends to provide better results, we gradually allow larger deviations and tune the $r$ by experiment, obtaining an appropriate setting as follows.

(1) For BigGAN, we only need to constrain the intermediate features:
\begin{itemize}
    \item Intermediate features: [2000, 2500, 3000, 3500, 4000, 4500, 5000, 5500, 6000]. 
\end{itemize}

(2) StyleGAN2 has more particularities we need to handle for feature domain optimization. In addition to the intermediate features, we optimize the noise vectors and apply the $l_1$ ball constraint to them at the same time. Involved in the generation of styles in StyleGAN2, the latent vectors also need to be optimized and we constrain their searching range within an $l_1$ ball as well:
\begin{itemize}
    \item Intermediate features: [2000, 3000, 4000, 5000]
    \item Noises: [1000, 2000, 3000, 4000, 5000]
    \item Latent vectors: [1000, 2000, 3000, 4000, 5000]
\end{itemize}

For the image fidelity regularization, we use $\alpha_ {TV} = 10^{-4}$, $\alpha_{l_ 2}=10^{-6}$. We run all experiments on NVIDIA RTX 2080 Ti GPUs and A100 GPUs. The experiments on the effects of K and on defense strategies are each conducted on 30 randomly selected images and the numerical results for batch size are the averages of 10 batches.

\end{document}